\documentclass{article}
\PassOptionsToPackage{dvipsnames}{xcolor}
\let\keptaddcontentsline\addcontentsline
\usepackage{iclr2027_conference,times}
\let\addcontentsline\keptaddcontentsline
\iclrfinalcopy
\usepackage{etoolbox}
\makeatletter
\patchcmd{\@maketitle}{Published as a conference paper at ICLR 2027}{}{}{}
\makeatother
\renewcommand{\headrulewidth}{0pt}
\renewenvironment{abstract}{\vskip.075in\centerline{\large\sc Abstract}\vspace{0.5ex}\par\noindent\ignorespaces}{\par\vskip 1ex}

\usepackage[utf8]{inputenc}
\usepackage[T1]{fontenc}
\usepackage{xspace}
\usepackage{graphicx,amsmath,amssymb,amsfonts,hyperref}
\usepackage{float}
\usepackage{xcolor}
\usepackage{booktabs}
\usepackage{mdframed}
\usepackage{caption}
\usepackage{needspace}
\usepackage{placeins}
\usepackage{microtype}
\usepackage{url}

\hypersetup{
  colorlinks=true,
  urlcolor=link,
  linkcolor=black,
  citecolor=link,
  filecolor=black,
  pdfborder={0 0 0}
}

\definecolor{link}{HTML}{0063BE}
\providecommand{\equationname}{Equation}
\providecommand{\sectionname}{Section}

\newcommand{\figref}[1]{%
  \figurename~\hyperref[#1]{\textcolor{link}{\ref*{#1}}}%
}
\newcommand{\tabref}[1]{%
  \tablename~\hyperref[#1]{\textcolor{link}{\ref*{#1}}}%
}
\newcommand{\eqrefc}[1]{%
  \equationname~\hyperref[#1]{\textcolor{link}{\ref*{#1}}}%
}
\newcommand{\secrefc}[1]{%
  \sectionname~\hyperref[#1]{\textcolor{link}{\ref*{#1}}}%
}

\newcommand{\github}{\raisebox{-1.3pt}{\includegraphics[height=1.05em]{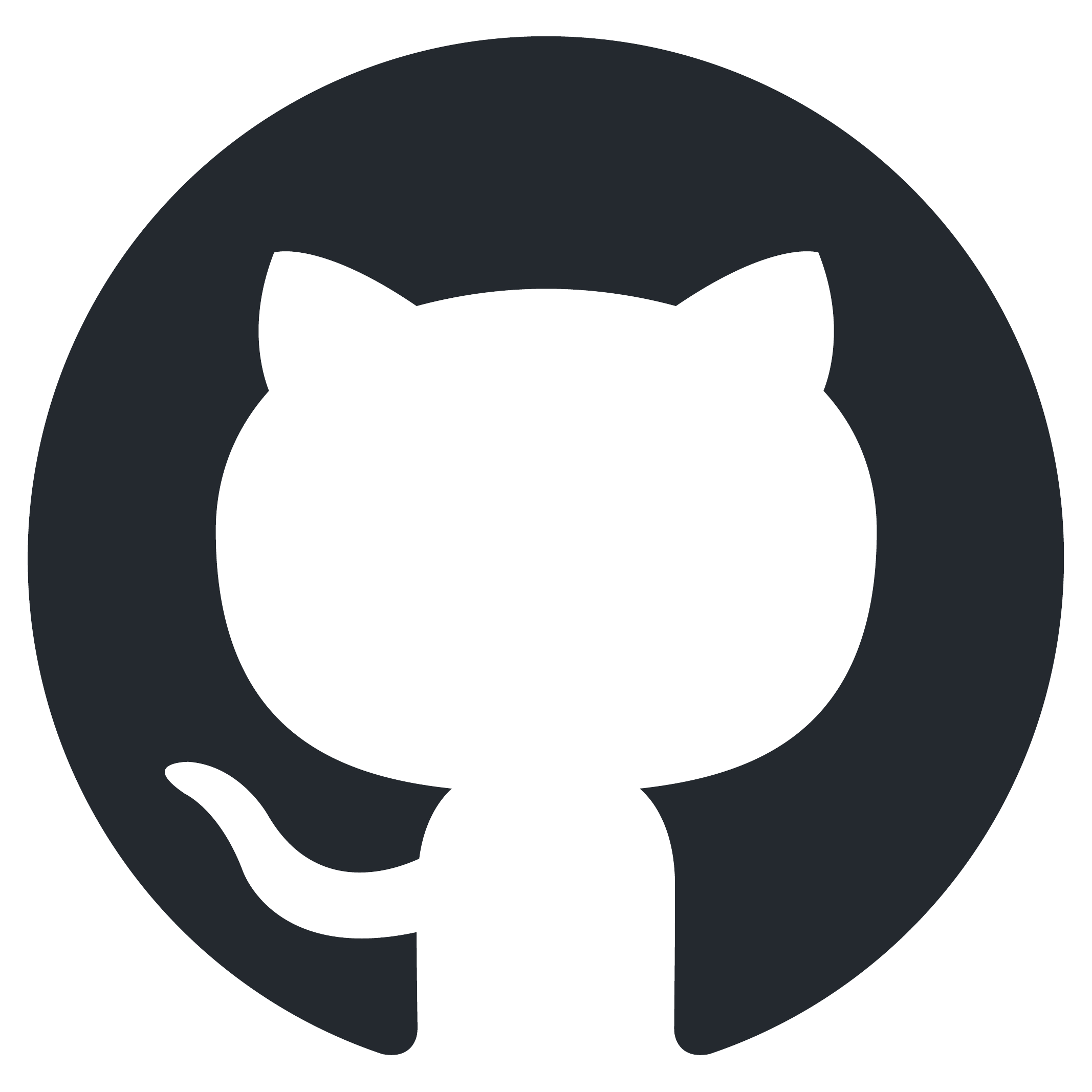}}\xspace}
\newcommand{\hf}{\raisebox{-1.3pt}{\includegraphics[height=1.05em]{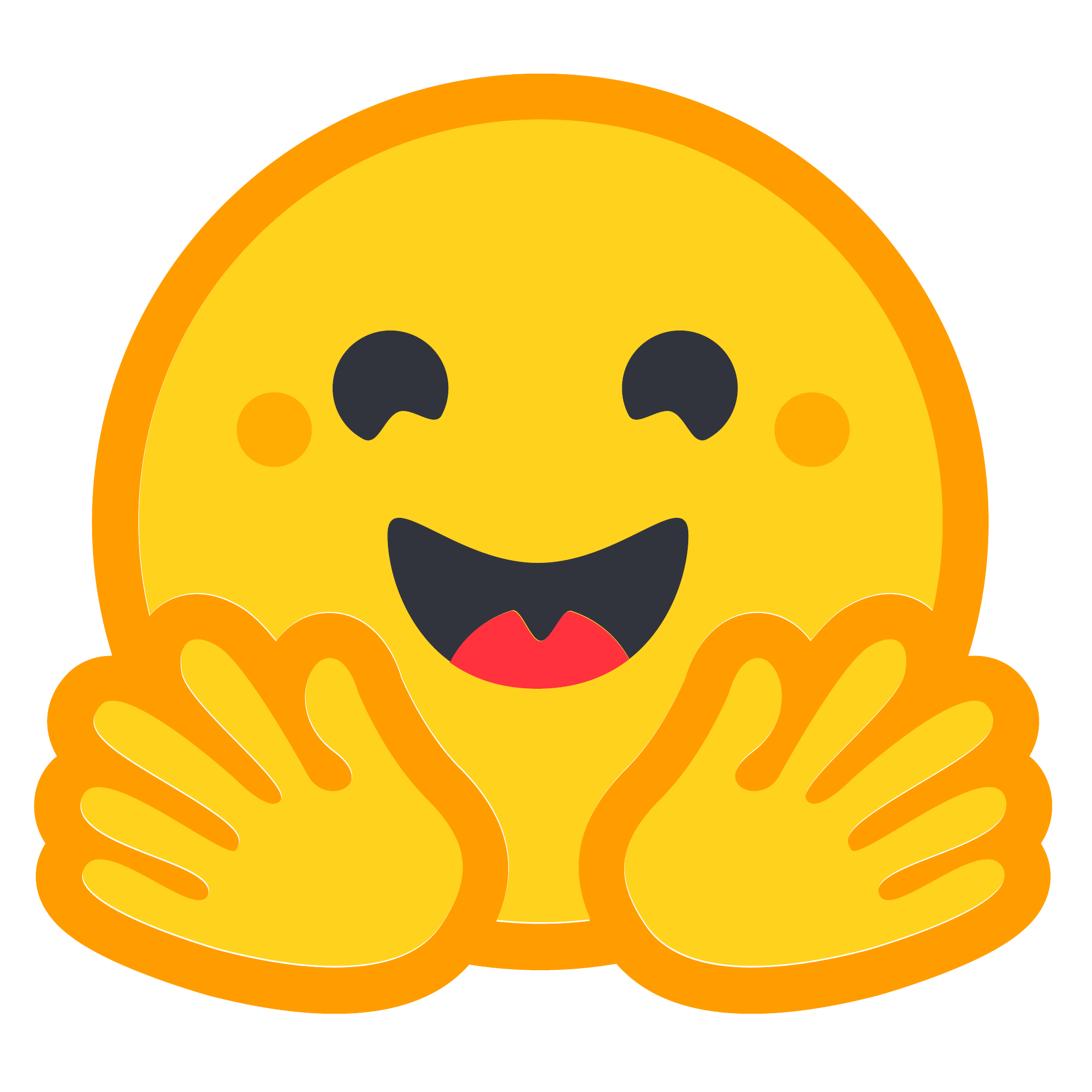}}\xspace}

\usepackage{bm}

\newenvironment{llmprompt}[1]{%
    \par\medskip
    \needspace{5\baselineskip}%
    \noindent\textbf{#1}\par\nopagebreak
    \begin{mdframed}[
        backgroundcolor=gray!5,
        roundcorner=2pt,
        linecolor=gray!40,
        leftmargin=0pt,
        innerleftmargin=10pt,
        innertopmargin=10pt,
        innerbottommargin=10pt,
        splittopskip=10pt,
        splitbottomskip=10pt,
        nobreak=false
    ]
    \setlength{\parindent}{0pt}
    \setlength{\parskip}{6pt}
    \small\ttfamily
}{%
    \end{mdframed}
}

\DeclareTextFontCommand{\textbf}{\bfseries}
\newcommand{\tightpar}{\looseness=-1 \spaceskip=\fontdimen2\font plus \fontdimen3\font minus 1.35\fontdimen4\font\relax}

\newcommand{\benchname}{\textsc{OccludeBench}}
\newcommand{\tracename}{\textsc{OccludeTrace}}

\title{\benchname{}: Can VLMs find what's missing?}

\author{Jasin Cekinmez\thanks{Equal Contribution} \quad
Addison J. Wu\footnotemark[1] \quad
Ryo Mitsuhashi\footnotemark[1] \\
\textbf{Linrong Cai} \quad
\textbf{Xingyu Fu} \quad
\textbf{Zhuang Liu}\\
Princeton University}

\begin{document}

\maketitle

\vspace{-0.5cm}
\begin{center}
    \small
    \hf\,\href{https://huggingface.co/datasets/afs07fda89sdfas90/data}{\textbf{Data}}
    \qquad
    \github\,\href{https://github.com/addisonwu05/OccludeBench}{\textbf{Code}}
\end{center}

\noindent\begin{minipage}{\linewidth}
    \centering
    \includegraphics[width=\linewidth]{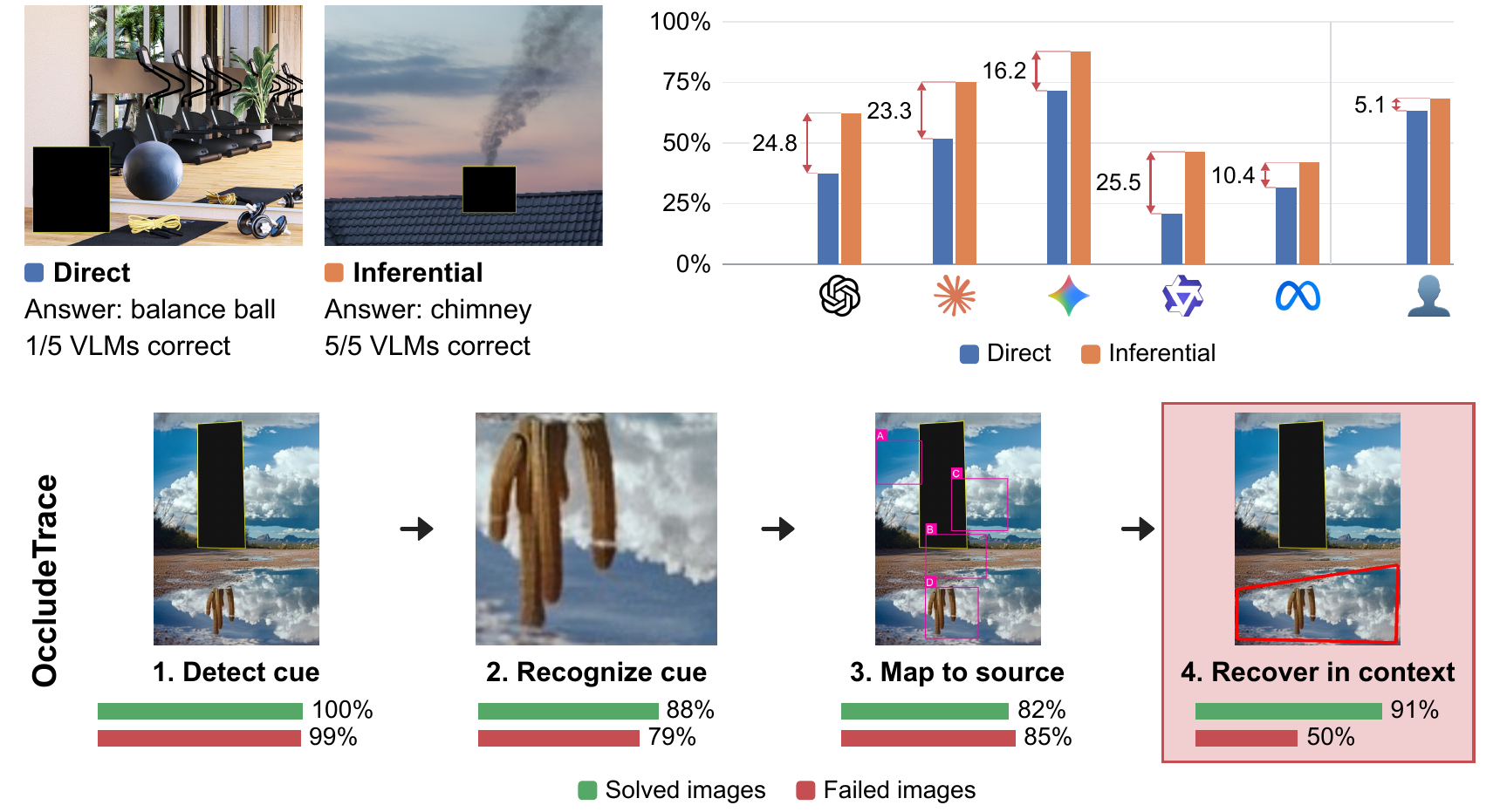}
    \captionsetup{type=figure}
    \caption{\textbf{Top:} \benchname{} asks VLMs to identify occluded objects from direct physical cues (shadows or reflections) or broader context. VLMs perform worse on direct cases than on inferential cases, while humans show no significant gap. \textbf{Bottom:} \tracename{} breaks recovery from a direct cue into four stages. Models generally succeed at detecting the cue, recognizing objects from a zoomed-in view, and matching objects to their projections. The main bottleneck is identifying the hidden object from its projection within the full image.}
    \label{fig:teaser}
\end{minipage}
\vspace{0.6em}

\newcommand{\abstractcontent}{%
How well can vision-language models (VLMs) reason about what is missing from an image? We study this question with \benchname{}, a manually curated benchmark for recovering fully occluded objects from the visual evidence that remains. \benchname{} distinguishes \textbf{direct} cases, where the hidden object must be recovered from a specific physical cue such as a shadow or reflection, from \textbf{inferential} cases, where it can be inferred from broader scene context. Across five VLMs, accuracy averages 42.7\% on direct cases versus 62.7\% on inferential cases, exhibiting a 20.0-point gap. Humans attain higher accuracies of 63.4\% versus 68.5\% for direct and inferential images, respectively, with no significant gap. To locate this failure, we introduce \tracename{}, a four-stage diagnostic that isolates where the visual reasoning process breaks down. The bottleneck emerges when models must recover the relevant object from the projection within the full image: they succeed on only 50\% of failed examples, compared with 91\% of examples they answer correctly. Extended reasoning, tool use, and few-shot prompting yield inconsistent improvements and do not reliably eliminate the gap. In contrast, an inpaint-then-confront intervention, which prompts models to first generate a candidate hidden object and then evaluate that hypothesis against the visible physical cue, reduces the gap to a statistically non-significant difference for both models with image-generation capabilities. Together, our results establish a tractable, controlled diagnostic for isolating how VLMs reason about visual content that is missing from an image.
}

\begin{abstract}
\abstractcontent
\end{abstract}

\addtocontents{toc}{\protect\setcounter{tocdepth}{-1}}
\section{Introduction}

\label{sec:intro}

Vision-language models (VLMs) have recently demonstrated strong performance across reasoning and vision-language benchmarks~\citep{lu2022learn, Yue_2024_CVPR, sarch2026vero}. Large-scale systems such as GPT and Gemini~\citep{achiam2023gpt, team2023gemini} further highlight the ability of these models to integrate information across modalities and solve complex tasks requiring both perception and reasoning. These successes concern content that is visible in the image. A basic question remains open: how well can VLMs reason about what is \emph{missing} from an image? Everyday scenes routinely leave objects out of view, and recovering them requires reasoning from the evidence that remains, whether a shadow, a reflection, or the surrounding context. Because most benchmarks use full, unconstrained images, it is difficult to determine which visual cues a model uses, and hence whether it reasons from the evidence or answers from prior expectations.
\begin{figure*}[h!]
    \includegraphics[width=\linewidth]{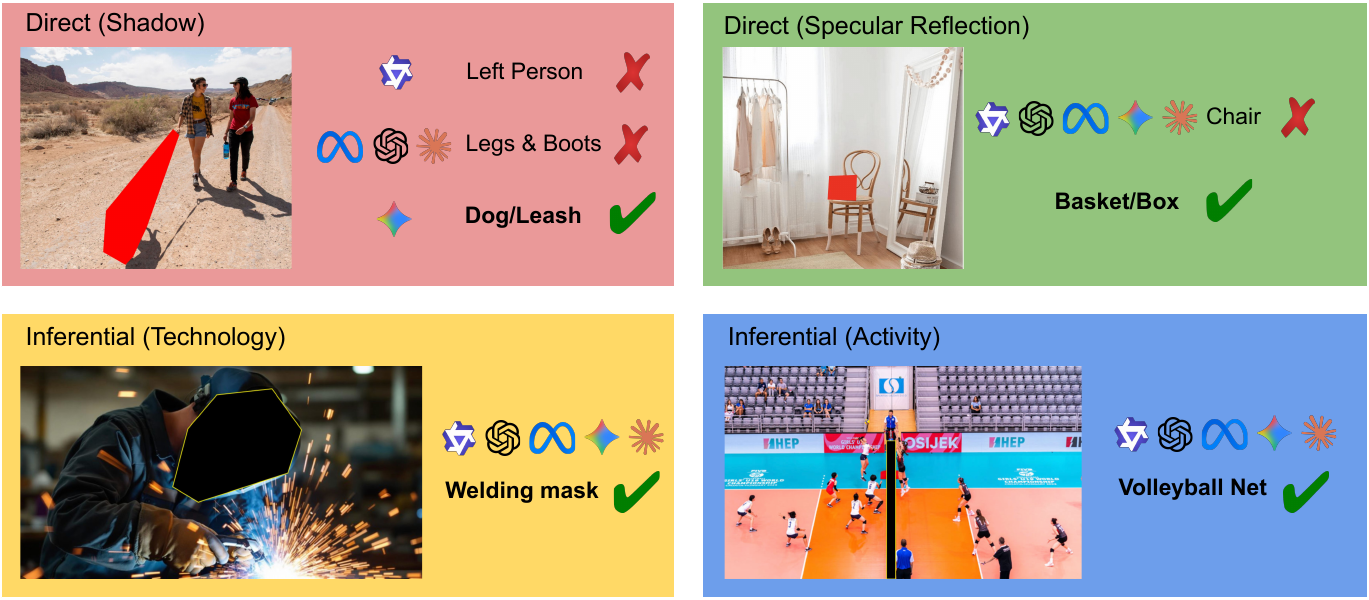}
    \caption{Examples of visual image problems involving direct or inferential evidence, along with their subcategories. We prompt the VLM to answer ``\textbf{What is the primary object that is completely occluded in this image? Be as specific as you can.}'' (see Appendix~\ref{app:basic_eval}). The responses of each model and the correct answer are shown on the right of each subfigure. Direct cases can be answered with a single visual cue, shadow or reflection, while inferential cases use broader context to arrive at the conclusion. In each question, we group the models by the response they each gave and indicate whether the answer is correct.}
    \label{fig:core_examples}
\end{figure*}

We study this question with \benchname{}, a manually curated benchmark for recovering fully occluded objects from the visual evidence that remains. We distinguish \textbf{direct} cases, where the hidden object must be recovered from a shadow or reflection, from \textbf{inferential} cases, where it can be inferred from broader scene context. Because both use the same task and masking procedure, the performance gap between them isolates differences in reasoning from specific physical evidence versus contextual inference (\figref{fig:core_examples}).

We find models perform worse on direct tasks than on inferential ones, while humans show no corresponding gap. To locate where the reasoning process breaks down, we introduce \tracename{}, a four-stage diagnostic that decomposes recovering a hidden object into detecting the projection, recognizing it in isolation, mapping it to its source, and recovering the object from the cue inside the full image. The bottleneck is not seeing the shadow or reflection, but isolating which object within it actually identifies what is hidden.

Attempts to mitigate this failure through in-context learning, tool use, extended reasoning, and targeted prompting yield inconsistent gains, and the gap between direct and inferential images persists. The exception is an inpaint-then-confront intervention: the model first generates a candidate hidden object and then evaluates that hypothesis against the visible physical cue, which reduces the gap to a statistically non-significant difference for both models with image-generation capabilities.

Our dataset and diagnosis suggest that evaluations can mask an important failure mode: models may answer correctly without using the visual signal that determines the answer. By contrasting cases where the hidden object can only be identified from a specific visual cue with cases where it can be inferred from broader context, we identify evidence selection as a key missing capability in current VLMs. More broadly, occlusion provides a tractable, controlled way to test how VLMs reason about content that is missing from an image, and \tracename{} shows how such a test can be decomposed to isolate where that reasoning fails.

\section{\benchname}
\label{sec:method}
\subsection{Dataset Creation}
\label{subsec:createdataset}
\begin{figure*}[t!]
    \includegraphics[width=0.95\textwidth]{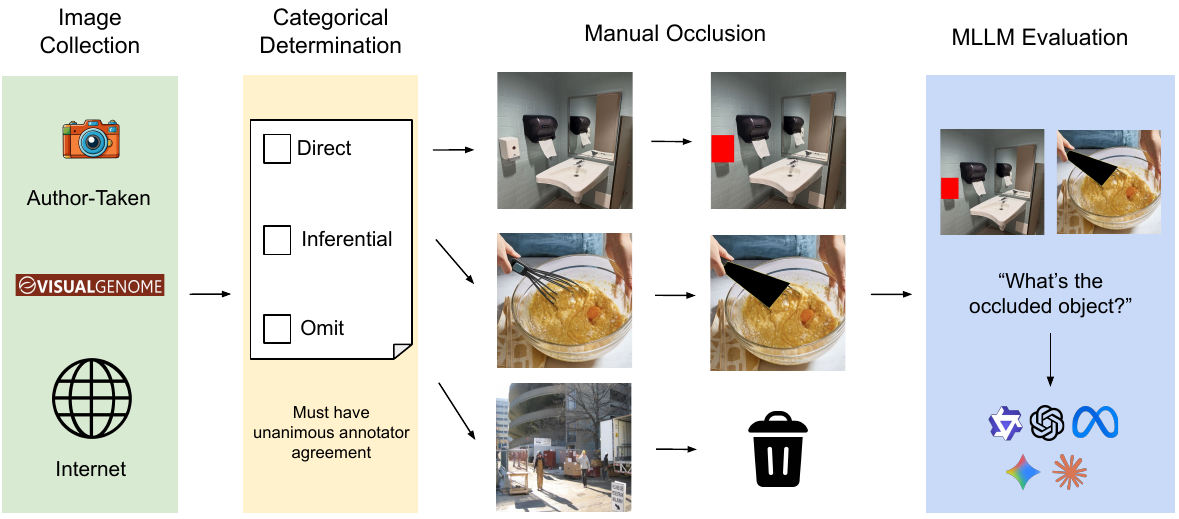}
    \caption{\looseness=-1 Overview of our dataset curation and evaluation pipeline. After we source images from the Internet, existing datasets, and self-collection, human annotators unanimously agree on the correct category for each image (\secrefc{subsec:createdataset}). We then evaluate how well the VLM identifies the object hidden in the manually occluded image.}
    \label{fig:pipeline}
\end{figure*}
\looseness=-1 \figref{fig:pipeline} provides an overview of the curation and evaluation pipeline used in this work. To create the dataset, we collected images from the Internet, Visual Genome~\citep{krishna2017visual}, and our own photographs, since synthetic images would weaken the ecological validity of the dataset. Initial testing also showed that AI-generated images for the direct category did not accurately portray the physical properties of shadows and reflections. Examples of relational object appearances generated by large language models (LLMs) were also repetitive, not diverse, and did not scale sufficiently. For the direct images, we selected examples that explicitly contain a shadow or reflection and where the occluded object cannot be reasonably inferred from context. For the inferential examples, we followed the same procedure but instead looked for diverse images in which the occluded object can be reasonably inferred from contextual cues. 

\paragraph{Dataset Distribution.}The direct category is composed of different types of physical projection cues, including specular reflections (47.7\%), diffuse reflections (28.4\%), and shadows (23.9\%). The inferential category spans a range of semantic domains, including technology (24.9\%), household scenes (24.3\%), culture (20.4\%), activities (17.1\%), and nature (13.3\%). \benchname{} contains 378 images, with 197 in the direct category and 181 in the inferential category.

\paragraph{Occlusion Procedure.} For each image, we artificially occlude a target object using a clearly visible bounding box or polygon, making it explicit that part of the scene is hidden through manual annotation; the bounding box or polygon coordinates are included in the dataset. We then include the image in our dataset if the human annotators unanimously agree that the image cleanly fits into the specified category. We use artificial occlusion rather than natural occlusion to avoid introducing confounds. For example, naturally occluded shadows or reflections may be distorted or missing. 
\subsection{Evaluation}
\paragraph{Task Setup.} {\tightpar We evaluate five state-of-the-art VLMs: Llama 4 Maverick~\citep{meta2025llama4}, Qwen 3 VL 8B~\citep{bai2025qwen3vl}, Claude Opus 4.6~\citep{anthropic2026opus46}, Gemini 3.1 Pro~\citep{googledeepmind2026gemini31pro}, and GPT-5.4~\citep{openai2026gpt54thinking}. We prompt the models for open-ended responses (not multiple-choice) on what they believe the occluded object to be. The task tests whether models can recover the hidden object from the remaining visual evidence, through direct cues or contextual understanding. We report accuracy based on whether the predicted object matches the ground truth under our judging framework.\par}

\paragraph{Grading.} We evaluate model predictions using an LLM-as-a-judge framework consisting of Claude Opus 4.6 and GPT-5.4 to determine whether the occluded object has been correctly identified~\citep{zheng2023judging}. We mark a prediction as correct if either judge deems it correct. We find that this LLM-as-a-judge framework achieves 98\% agreement with ground-truth annotations by the authors, slightly higher than the agreement rate of single-judge evaluations, indicating high reliability. Because judging only checks whether a free-form answer names the ground-truth object, it leaves little room for self-preference bias from judges that are also evaluated models. All prompts for evaluation and LLM-as-a-judge can be found in Appendix~\ref{app:prompts}.

\subsection{Human Baseline}

We collect a human baseline using Prolific~\citep{prolific}, recruiting 473 English-speaking participants from the United States, United Kingdom, Canada, Australia, and New Zealand. \textbf{Each participant is presented with 6 questions, consisting of 3 direct and 3 inferential items.} Questions are randomly sampled from our dataset under the constraint that all questions are approximately equally represented across participants. The order of the 6 questions is independently randomized for each participant, and questions are shown sequentially, one at a time. Responses are collected in a free-form setting without multiple-choice options.

\paragraph{Compensation and Protocol.} Participants are compensated at \$12.50/hour, with a median completion time of 2.5 minutes. The task is zero-shot: participants do not see example questions.

\section{Results}

\label{sec:results}
\subsection{Main Results}
\paragraph{VLMs struggle more with identifying occluded objects in direct images than in inferential ones.} VLMs identify the occluded object correctly in 52.3\% of cases on average (\tabref{tab:merged_results}). Gemini 3.1 Pro has the highest total accuracy at 79.4\% among all inferential and direct images, while Qwen 3 VL 8B attains the lowest at 33.2\%. However, across all five VLMs evaluated, accuracy is highly variable between direct and inferential images; VLMs consistently attain lower accuracies when discerning the occluded object for direct images (20.0 p.p. lower on average) than for inferential images. While the smallest gap is found in Llama 4 Maverick at 10.4 p.p., the rest of the VLMs have performance deficiencies of at least 16.2 p.p., with the highest gap being observed with Qwen 3 VL 8B at 25.5 p.p. Lastly, performance is also largely uniform across occlusion cue categories (\figref{fig:category_accuracy}, Appendix~\ref{app:category_uniformity}).

\begin{figure}
    \centering
    \includegraphics[width=0.95\textwidth]{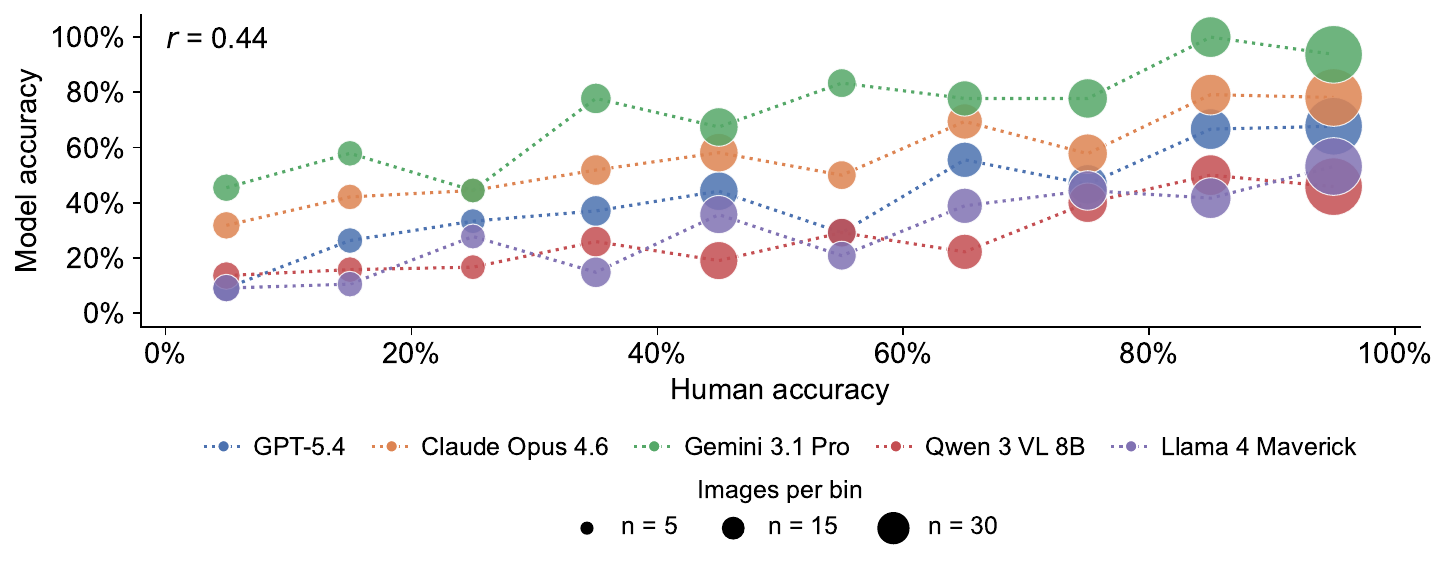}
    \caption{Model versus human accuracy for occluded object identification. Each dot is a 10-point bin of per-image human accuracy, sized by the number of images. Across all VLMs, questions that are harder for humans are also harder for models ($r = 0.44$ between per-image human accuracy and mean model accuracy).}
    \label{fig:human_model_accuracy}
\end{figure}

\paragraph{Most VLMs perform worse than humans at identifying the occluded object.}
Across all individual responses, humans achieve an accuracy of 65.9\%, exceeding four of the five evaluated VLMs. Moreover, unlike VLMs, humans do not exhibit a statistically significant performance gap between direct and inferential images: human accuracy is 63.4\% on direct images and 68.5\% on inferential images, compared with model averages of 42.7\% and 62.7\%, respectively (Table~\ref{tab:merged_results}). Images that humans find difficult also tend to be difficult for models ($r = 0.44$), in both the direct and inferential categories (\figref{fig:human_model_accuracy}, Appendix~\ref{app:difficulty-distribution}).

\begin{table}[t]
\centering
\renewcommand{\arraystretch}{1.15}
\setlength{\abovecaptionskip}{8pt}
\small
\begin{tabular}{lccc}
\toprule
\textbf{Model} & \textbf{Direct (\%)} & \textbf{Inferential (\%)} & \textbf{Gap} \\
\midrule
GPT-5.4           & 37.6 & 62.4 & 24.8 \\
Claude Opus 4.6   & 51.8 & 75.1 & 23.3 \\
Gemini 3.1 Pro    & 71.6 & 87.8 & 16.2 \\
Qwen 3 VL 8B      & 20.9 & 46.4 & 25.5 \\
Llama 4 Maverick  & 31.6 & 42.0 & 10.4 \\
\bottomrule
\end{tabular}
\caption{VLM accuracy on occluded object identification, split by direct and inferential images. Gap is inferential minus direct accuracy in percentage points. Every model scores lower on direct images.}
\label{tab:merged_results}
\end{table}

\subsection{\tracename{}: Why Do VLMs Fail?}
\label{subsec:occludetrace}
\begin{figure}[t]
    \centering
    \includegraphics[width=0.95\textwidth]{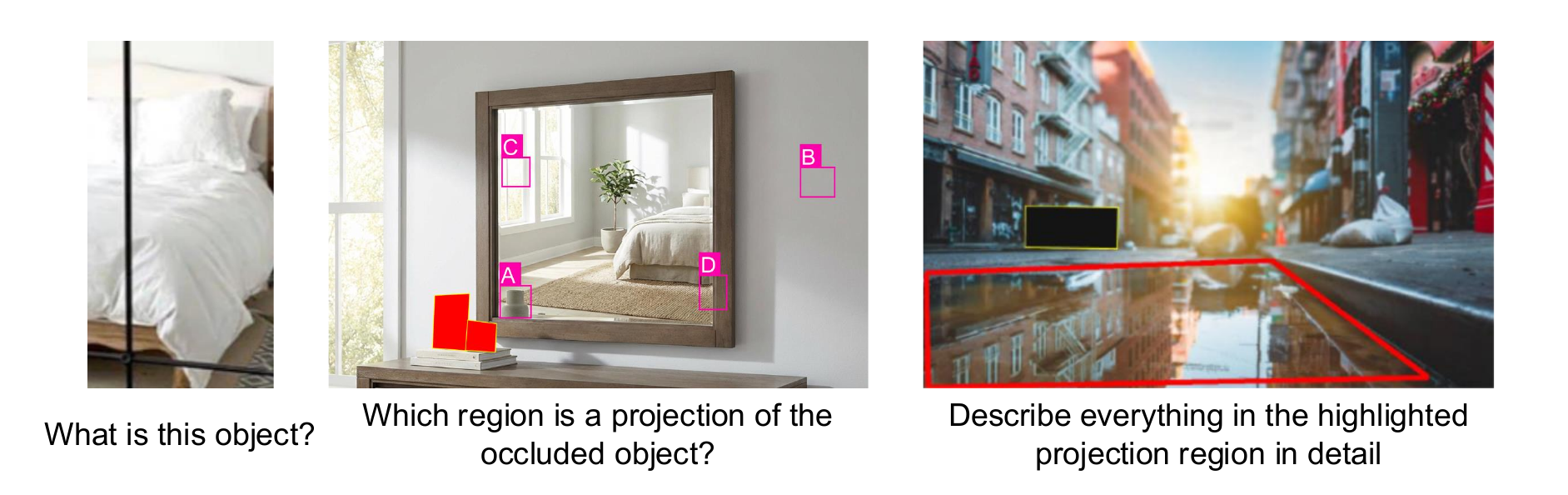}
    \caption{Examples of the \tracename{} stages. \textbf{(Left)} Zoomed-in object recognition: the model sees only the cropped projection and names the object. \textbf{(Middle)} Source mapping: the model chooses which of four candidate regions is the source of the marked projection. \textbf{(Right)} Highlighted-region recovery: the model describes everything in the marked projection region, with no mention of shadows, reflections, or occlusion.}
    \label{fig:stage_breakdown}
\end{figure}

VLMs misidentify occluded objects in direct images more frequently; we now examine where this failure occurs. Identifying an object from its shadow or reflection requires several skills in sequence: noticing that a projection exists, reading the object it depicts, linking the projection to the part of the scene that casts it, and recovering the object from the projection while it sits inside the full image. Any one of these could be the point of failure. We therefore develop \tracename{}, a four-stage diagnostic that isolates each skill with its own prompt on the same direct images (examples in \figref{fig:stage_breakdown}). A stage that causes the failures should show a large gap between the groups, whereas a stage that models handle equally well in both groups is not the bottleneck. We describe each stage and its results (\tabref{tab:stage_accuracy}).

\begin{table}[t]
\centering
\small
\renewcommand{\arraystretch}{1.15}
\setlength{\abovecaptionskip}{10pt}
\begin{tabular}{lcc}
\toprule
\textbf{\tracename{} stage} & \textbf{Accuracy on failed examples} & \textbf{Accuracy on successful examples} \\
\midrule
1. Cue detection                & 99\% & 100\% \\
2. Zoomed-in object recognition & 79\% & 88\%  \\
3. Source mapping               & 85\% & 82\%  \\
4. Highlighted-region recovery  & 50\% & 91\%  \\
\bottomrule
\end{tabular}
\caption{Accuracy at each \tracename{} stage on direct images, computed separately over the examples answered incorrectly and correctly in the base condition. Only the final stage separates the two groups.}
\label{tab:stage_accuracy}
\end{table}

\paragraph{Stage 1: Cue detection.} We ask whether the image contains a shadow or reflection. Models answer correctly on 99\% of failed examples and 100\% of successful ones, so models notice the projection even when they go on to answer incorrectly.

\paragraph{Stage 2: Zoomed-in object recognition.} We crop the projection out of the image and ask the model what object it shows. Accuracy is 79\% on failed examples and 88\% on successful ones. When the projection is presented in isolation, models read its content about equally well regardless of whether they later fail.

\paragraph{Stage 3: Source mapping.} {\tightpar We mark the projection with a transparent bounding box, occlude four candidate regions, and ask which region casts the shadow or reflection. Models choose the correct region on 85\% of failed and 82\% of successful examples, so they can link the projection to its source.\par}

\paragraph{Stage 4: Highlighted-region recovery.} We mark the projection with a transparent bounding box in the intact image and ask the model to describe everything in the region in detail (prompt in Appendix~\ref{app:behavior}). We then check whether the description names the depicted object. This is the only stage where the two groups diverge: models recover the object on 91\% of successful examples but on only 50\% of failed ones.

\paragraph{Diagnosis.} The first three stages rule out the obvious explanations. Failures are not caused by missing the cue, by being unable to read a shadow or reflection, or by being unable to link it to its source. Models fail at the stage where the projection must be read while embedded in the complete image, even when the region is pointed out to them. The bottleneck is visual extraction in context rather than reasoning about physical cues, which motivates the interventions in \secrefc{subsec:improve}.

\subsection{Can We Close the Gap?}
\label{subsec:improve}
\begin{figure}[t!]
    \centering
    \includegraphics[width=0.95\textwidth]{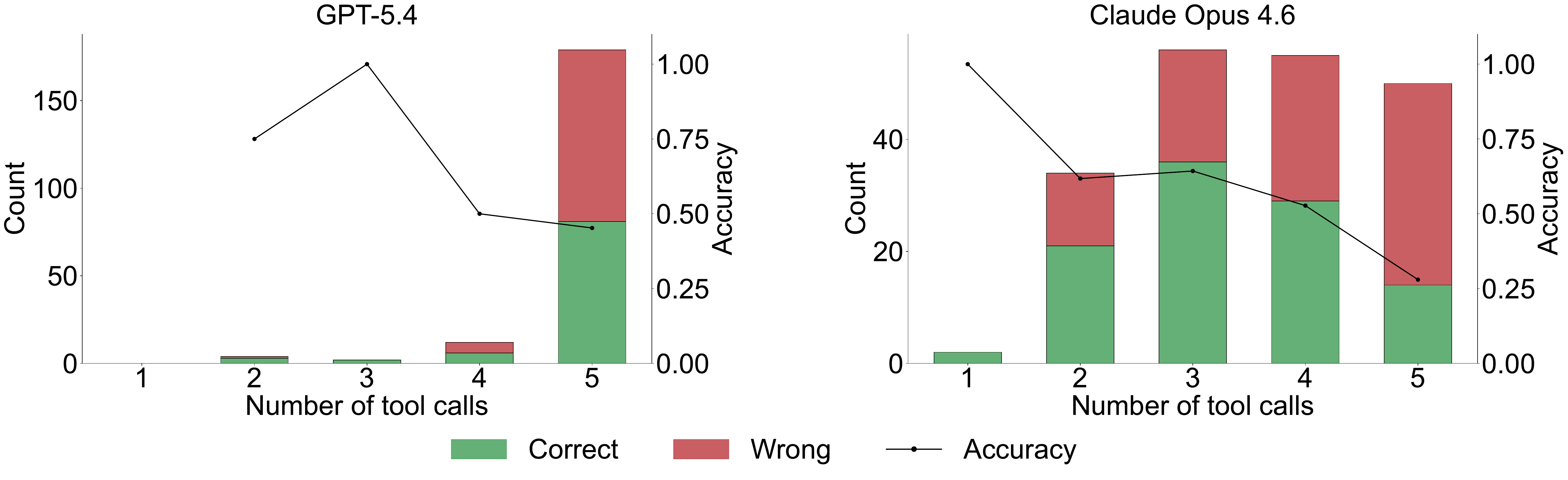}
    \vspace{0.5em}
    \caption{\looseness=-1 Tool-call analysis on direct images. \textbf{Left:} GPT-5.4 accuracy by number of tool calls. GPT-5.4 exhausts its 5-call budget ($n=179$) at 45\% accuracy. There were no traces where GPT-5.4 used exactly one tool. \textbf{Right:} Claude Opus 4.6 accuracy by number of tool calls; accuracy declines monotonically from 62\% (2 calls) to 28\% (5 calls).}
    \label{fig:tool_call_analysis}
\end{figure}

\begin{figure}
    \includegraphics[width=0.90\textwidth]{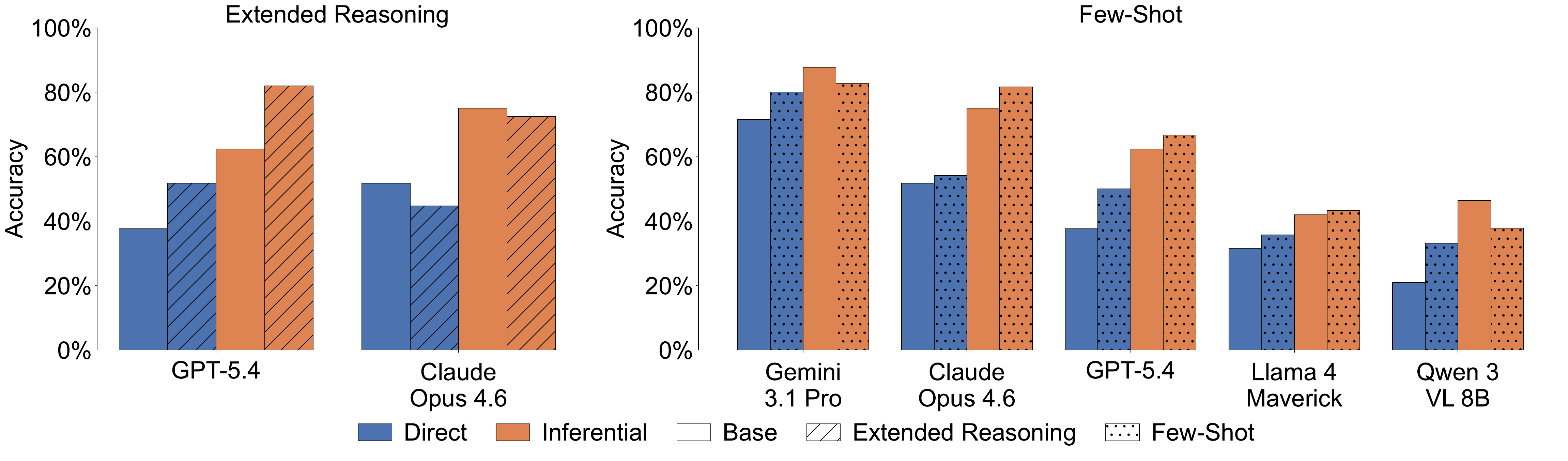}
    \caption{\tightpar Accuracies when attempting to improve VLM performance on the occluded object task. Extended reasoning and few-shot prompting improve performance on direct images for most models, but do not reliably close the performance gap between direct and inferential images. Inferential performance sometimes decreases.}
    \label{fig:improvements}
\end{figure}

\paragraph{Extended Reasoning.} We investigate whether additional reasoning can improve performance overall. We evaluate the performance of GPT-5.4 and Claude Opus 4.6 with their respective highest reasoning modes activated on the same initial experimental setup. As shown in \figref{fig:improvements}, Claude Opus 4.6 decreases in accuracy for both direct and inferential images, whereas GPT-5.4 has accuracy increases for both categories. However, both GPT-5.4 and Claude Opus 4.6 still perform worse on direct than on inferential images.

\paragraph{Tool Use.} We investigate whether augmenting VLMs with image-based tools can aid performance on direct images. We use tools commonly used in image manipulation, including cropping, reflecting, edge detection, rotating, and zooming (see Appendix~\ref{app:tools} for the details). We test GPT-5.4 and Claude Opus 4.6 with high reasoning. GPT-5.4 decreases in accuracy from 51.8\% to 46.7\%. Tool usage does not manage to bring the performance of Claude Opus 4.6 on direct images (45 to 52\%) up to that for inferential images. More tool calls do not translate to higher accuracy: for both models, the correct-answer group uses \emph{fewer} tools on average than the incorrect group. The two models also exhibit very different tool-call profiles (\figref{fig:tool_call_analysis}), with GPT-5.4 nearly always exhausting its 5-call budget and relying heavily on \texttt{crop} and \texttt{zoom}, while Claude Opus 4.6 distributes calls more evenly and rarely crops. Expanding to full agentic harnesses with unrestricted code execution and high reasoning does not close the gap either. Codex with GPT-5.4 reaches 51.8\% and Claude Code with Claude Opus 4.6 reaches 34.5\%, both still well below inferential performance.

\begin{figure*}[t]
    \includegraphics[width=\linewidth]{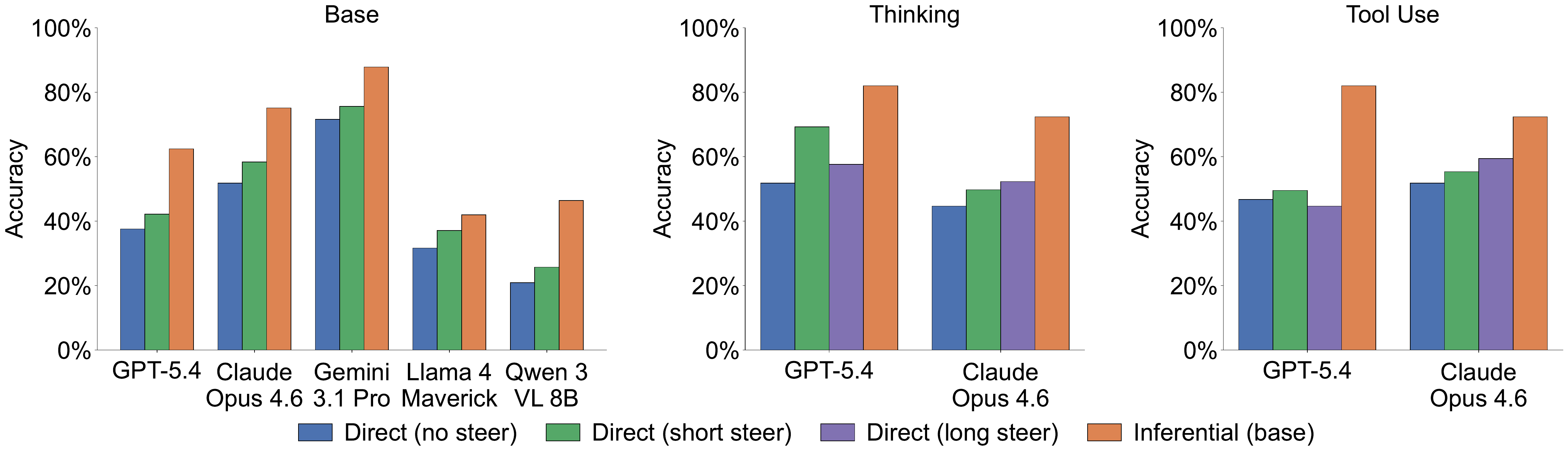}
    \caption{\looseness=-1 Accuracy on occluded object identification. All VLMs perform worse on direct images than on inferential ones. Tool use and prompt steers yield mixed improvements that vary across models and are insufficient to close the gap.}
    \label{fig:core_accuracy}
\end{figure*}

\paragraph{Steering Prompts.} We additionally test two steering prompts that direct models to attend to projections: a short steering prompt (``Pay close attention to any reflective surfaces or shadows in the image as you will need it to answer the question.'') and a longer steering prompt that further specifies what counts as a valid answer (full text in Appendix~\ref{app:nudges}). As shown in \figref{fig:core_accuracy}, both steers yield consistent improvements across base, thinking, and tool-use settings, with the largest gains on Claude Opus 4.6 + tool use (51.8\% $\rightarrow$ 59.4\%) and GPT-5.4 + thinking (51.8\% $\rightarrow$ 69.2\%). However, no steer configuration closes the gap to inferential performance, suggesting models can be cued to consider physical projections like shadows and reflections but still fail to reason about them reliably.

\paragraph{Few-Shot Prompting.} \looseness=-1 We find that few-shot prompting does not consistently reduce the performance gap across models. For the few-shot experiment, we include one direct and one inferential example (see Appendix~\ref{app:improvement}). We hold out these two examples from the evaluation. As shown in \figref{fig:improvements}, while Gemini 3.1 Pro and Qwen 3 VL 8B have small performance gaps between direct and inferential images (80 vs. 83\% and 33 vs. 38\%, respectively), gaps still exist for GPT-5.4 (50 vs. 67\%), Claude Opus 4.6 (54 vs. 82\%), and Llama 4 Maverick (36 vs. 43\%). Furthermore, we observe that the smaller gap with Qwen 3 VL 8B is associated with reduced accuracy for inferential images (46.4\% to 37.8\%).

\begin{figure}[t]
    \centering
    \includegraphics[width=\linewidth]{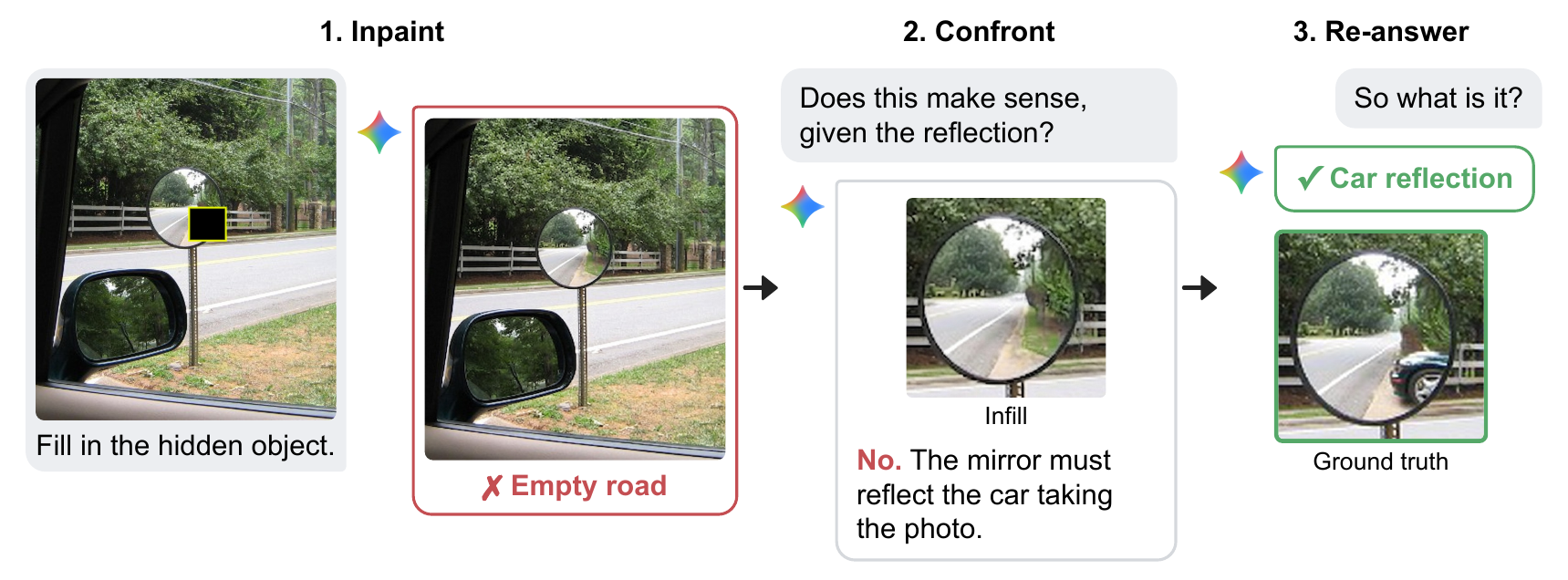}
    \caption{Inpaint-then-confront turns identification into hypothesis testing. Reading a hidden object directly off an indirect cue is hard, but checking whether a proposed object is consistent with that cue is easier. \textbf{(1)} Asked to fill in the masked part of a roadside convex mirror, Gemini 3.1 Pro paints in only the empty road. \textbf{(2)} Shown the original next to its infill, the model rejects the infill: the photo is taken from inside a car facing the mirror, so the mirror must reflect that car. \textbf{(3)} It then identifies the hidden object as the car's reflection, which it missed at baseline. For readability, the figure splits the single confront prompt into its two questions, and the bubbles paraphrase the model's responses.}
    \label{fig:inpainting_example}
\end{figure}

\paragraph{Inpainting Multi-step Execution.}
We next test whether models can overcome this gap by converting open-ended object identification into a more discriminative, iterative task. We first ask the model to inpaint the occluded region with the hidden object made visible (prompts in Appendix~\ref{app:inpainting}). We then present the resulting image and ask, ``Does this infill make sense physically, given the properties of reflections and shadows? Now, what is the occluded object? Be as specific as you can.'' The generated infill provides a concrete hypothesis for the model to evaluate (\figref{fig:inpainting_example}): mismatches between the proposed object and the observed shadow or reflection expose constraints on what the hidden object could instead be. Thus, rather than identifying the object directly from an indirect physical cue, the model can discriminate among and revise explicit visual hypotheses.

\looseness=-1 This is the most effective intervention we evaluate. Extended reasoning and tool use fail to eliminate the significant gap for either tested model, while few-shot prompting does not do so consistently across models. In contrast, inpaint-then-confront eliminates the significant gap for both GPT-5.4 and Gemini 3.1 Pro, the two models with image-generation capabilities (\tabref{tab:model_comparison_combined}). This suggests that direct physical evidence is not simply inaccessible to these models: once the inference is externalized as a visual hypothesis that can be inspected and rejected, they are substantially better able to detect and utilize it.

\begin{table}[t]
\centering
\small
\renewcommand{\arraystretch}{1.15}
\setlength{\abovecaptionskip}{10pt}
\begin{tabular}{llccc}
\toprule
\textbf{Model} & \textbf{Comparison} & \textbf{Rates} & \textbf{Difference} & \textbf{$\bm{p}$} \\
\midrule
GPT-5.4 & baseline direct vs inferential & 37.6\% vs 62.4\% & +24.8pp & <0.0001 \\
GPT-5.4 & inpaint direct vs inferential   & 54.3\% vs 62.4\% & +8.1pp  & \textbf{0.12} \\
\midrule
Gemini 3.1 Pro & baseline direct vs inferential & 71.6\% vs 87.8\% & +16.2pp & 0.0001 \\
Gemini 3.1 Pro & inpaint direct vs inferential   & 82.2\% vs 87.8\% & +5.6pp  & \textbf{0.15} \\
\bottomrule
\end{tabular}
\caption{Accuracy of inpainting multi-step evaluation for GPT-5.4 and Gemini 3.1 Pro. This intervention largely eliminates the gap between identifying the missing object across direct and inferential settings.}
\label{tab:model_comparison_combined}
\end{table}

\section{Related Work}
\label{sec:relatedwork}

\paragraph{VLMs and Evaluations.}

VLMs have advanced rapidly through integration of vision encoders with LLM backbones~\citep{FLAMENGO, liu2023llava, dai2023instructblip}, with recent unified models also supporting generation~\citep{xie2025showo, Wu_2025_CVPR, wu2025vilau}. They achieve strong performance on visual understanding benchmarks, from perception and general multimodal evaluation~\citep{VQA, Li_2024_CVPR, mmbench, MME} to complex reasoning~\citep{lu2024mathvista, Yue_2024_CVPR, mmvet}. Yet vision-centric reasoning is far from solved~\citep{visulogic}, and reasoning about hidden objects remains underexplored.

\paragraph{Occlusion and Indirect Visual Evidence.}
\looseness=-1 Amodal completion, inferring occluded objects via semantics or symmetry, is a recognized core visual reasoning ability~\citep{Zhan_2024_CVPR, Ao_2025_CVPR, Pothiraj_2025_ICCV}. While recent benchmarks probe spatial reasoning~\citep{visual-spatial, blink} and partial occlusion~\citep{liu2025visiblebenchmarkingocclusionperception}, cases of total occlusion remain underexplored. Physical commonsense benchmarks~\citep{physion, chow2025physbench} test related intuitions, but they rarely require models to reason from indirect visual cues like projections or reflections. Where prior work treats shadows and reflections as noise to be removed~\citep{shadow-removal, removing-reflection, hu2026unveiling}, we use these indirect cues as the primary evidence for reasoning about fully hidden objects. 

\paragraph{Inference and Evaluation Gaps.} 
Identifying occluded objects also draws on commonsense and world knowledge, the focus of commonsense and knowledge-based VQA benchmarks~\citep{Marino_2019_CVPR, commonsense, aokvqa} and methods~\citep{hind}. Humans use scene grammar to guide attention and perception~\citep{scene-grammar} and perceive partly occluded objects as unified wholes by four months of age~\citep{KELLMAN1983483}, but whether VLMs use such cues remains unclear. Recent work probes VLM perception of out-of-context objects~\citep{oodbench}, whereas we study objects that are completely hidden. Meanwhile, broader scrutiny of VLM benchmarks reveals that many can be solved without visual input~\citep{chen2024are}, models fail on low-level visual tasks where humans succeed~\citep{blink, Tong_2024_CVPR}, and task-relevant visual information captured by encoders often goes unused by downstream modules~\citep{fu2025hidden, liu2026seeing}.

\section{Conclusion and Discussion}
\label{sec:discussion}

\benchname{} asks how well VLMs reason about what is missing from an image, using occlusion as a controlled diagnostic: we choose what is hidden, know the answer, and control the evidence that remains. The answer is that VLMs often fail not because they cannot perceive visual information, but because they do not reliably extract and use the evidence required to recover the missing object. VLMs achieve only 52.3\% accuracy compared to 65.9\% for humans, with a 20.0-percentage-point drop on direct evidence cases, a gap not observed in human performance. \tracename{}, our four-stage diagnostic, isolates where the reasoning process breaks down: models detect that a cue exists, recognize it in isolation, and map it to its source, but struggle to detect the relevant object within the projection once it is embedded in the full image. The most effective intervention we tested, inpaint-then-confront, has the model generate a candidate hidden object and then evaluate that hypothesis against the visible physical cue; it reduces the gap to a statistically non-significant difference for both models with image-generation capabilities, suggesting that making a hypothesis concrete enough to check against the evidence helps models catch their own mistakes.

\benchname{} contains 378 manually curated images. Its simple design around reflections and shadows lets us isolate whether VLMs can reliably reason from exact visual evidence in a setting straightforward for humans. Our results suggest that current evaluations can overestimate visual reasoning by allowing models to succeed through partial cues or learned associations rather than the evidence that determines the answer. More broadly, these findings establish occlusion as a tractable, controlled diagnostic for how VLMs reason about missing visual content, highlight reliable use of task-relevant evidence as an open challenge for current VLMs, and motivate future systems that more faithfully ground predictions in visual input.

\section*{Acknowledgments}
{\hyphenpenalty=10000 This research was supported by Princeton University's Office of Undergraduate Research Undergraduate Fund for Academic Conferences.\par}

\bibliographystyle{iclr2027_conference}
\bibliography{main}

\newpage
\section*{\Large Appendix}

\vspace{.2cm}

\appendix

\addtocontents{toc}{\protect\setcounter{tocdepth}{2}}
\renewcommand{\contentsname}{Contents}
{\hypersetup{linkcolor=black}\setlength{\parskip}{0pt}\tableofcontents}
\vspace{.3cm}

\section{Prompts}
\label{app:prompts}

This appendix collects the full text of every prompt used in our evaluation pipeline. Each subsection corresponds to the section of the main body where the prompts are used, and we briefly describe the role of each prompt before listing it verbatim. All prompts are issued at temperature~0 with open-ended (free-form) responses; we do not use multiple-choice options at any stage.

\subsection{Evaluation}
\label{app:basic_eval}

\paragraph{Pipeline Overview.} The base evaluation issues the query below and records the model's response.
\begin{llmprompt}{Main Occluded Query}
What is the primary object that is completely occluded in this image? Be as specific as you can.
\end{llmprompt}

\subsection{Judging}
\label{app:judging}
\paragraph{Ensemble Decision Rule.} We use Claude Opus 4.6 and GPT-5.4 as independent judges. Each judge receives the ground-truth label and the model's free-form response, and returns a binary correctness verdict. Across the 378 items, this two-judge ensemble agrees with author-provided gold labels 98\% of the time, exceeding either single-judge agreement rate.

\begin{llmprompt}{Judge Correctness (Model Response)}
You are evaluating whether a vision model correctly identified an occluded object.

Ground truth: "\{ground\_truth\}"

Model response: "\{response\}"

Does the model's response correctly identify the same object as the ground truth?
Allow for reasonable variation in phrasing, specificity, or synonyms
(e.g., `barber pole' matches `barber's pole', `wing mirror' matches `side mirror',
`cat' matches `pet' if the ground truth is `Reasonable Pet').
The response should identify the core object — partial credit guesses that get the
category right but miss specifics should be marked correct only if the category-level
answer is what the ground truth describes.
Answer only `yes' or `no'.
\end{llmprompt}

\subsection{Robustness Experiments}

\paragraph{Localization Control.} The transparent-box query is used in the unmasked-bounding-box experiment to rule out localization failures (\tabref{tab:unmasked_results}). The image is shown with the bounding box drawn as a yellow outline and the underlying contents visible, and the model is asked what the box contains. Asking about the box rather than about ``what is occluded'' keeps the prompt symmetric across direct and inferential items and avoids leaking the answer's modality.

\paragraph{Localization is not the bottleneck.} Another concern is that VLMs may fail to identify the location of the occluded object in the first place. To address this, we evaluate how well VLMs identify the object within the occlusion bounding box, but unmasked. This serves as a proxy test for identifying the location of the occlusion bounding box itself. VLMs do so 89\% to 93\% of the time, with near-equal performance for both direct (86\% to 94\% accuracy) and inferential (90\% to 93\% accuracy) images (\tabref{tab:unmasked_results}). Higher accuracy on transparent than on occluded recognition suggests that failures cannot be primarily attributed to difficulties in localizing the occluded region, so identifying the bounding box is not a confound.

\begin{table}[h]
\centering
\renewcommand{\arraystretch}{1.15}
\setlength{\abovecaptionskip}{8pt}
\small
\begin{tabular}{lcccc}
\toprule
& \multicolumn{2}{c}{\textbf{Occluded (\%)}} & \multicolumn{2}{c}{\textbf{Unmasked (\%)}} \\
\cmidrule(lr){2-3} \cmidrule(lr){4-5}
\textbf{Model} & Direct & Infer. & Direct & Infer. \\
\midrule
GPT-5.4           & 37.6 & 62.4 & 91.9 & 93.4 \\
Claude Opus 4.6   & 51.8 & 75.1 & 86.3 & 92.8 \\
Gemini 3.1 Pro    & 71.6 & 87.8 & 94.4 & 92.3 \\
Qwen 3 VL 8B      & 20.9 & 46.4 & 89.3 & 90.1 \\
Llama 4 Maverick  & 31.6 & 42.0 & 87.8 & 89.5 \\
\bottomrule
\end{tabular}
\caption{VLM accuracy on occluded and unmasked images. \textit{Occluded}: original images with the object hidden. \textit{Unmasked}: bounding box shown but contents visible.}
\label{tab:unmasked_results}
\end{table}

\begin{llmprompt}{Transparent-Box Query}
What is the object outlined by the yellow border in this image? Be as specific as you can.
\end{llmprompt}

\subsection{\tracename{} Prompts}
\label{app:behavior}

\begin{llmprompt}{Stage 1: Cue Detection}
Is there the presence of a reflection or a shadow in this image?
\end{llmprompt}

\begin{llmprompt}{Stage 2: Zoomed-In Object Recognition}
Depicted in this image is a shadow or reflection of another object. What is the original object that it is a shadow or reflection of? Be as specific as you can.
\end{llmprompt}

\begin{llmprompt}{Stage 3: Source Mapping}
The shadow or reflection of an object is clearly visible in this image. One of the four labeled regions (A, B, C, D) is occluded (blacked out) and hides the actual object that casts that shadow or reflection. Which labeled region contains the hidden object? Answer with only the single letter: A, B, C, or D.
\end{llmprompt}

\paragraph{Probing Scene Understanding Without Cue Framing.} The Comprehensive Scene Description prompt is used in the projection-region probing experiment (\secrefc{subsec:occludetrace}, Stage 4). To separate perception failures from reasoning failures, the prompt avoids any mention of shadows, reflections, or occlusion and just asks for an exhaustive description of the highlighted region. We then check post hoc whether the description names the object depicted within the projection.

\begin{llmprompt}{Comprehensive Scene Description of Highlighted Region}
This image contains a thick-bordered transparent bounding box (the border may be yellow, red, or green)
highlighting a region of the scene. Provide a comprehensive, detailed description of the entire scene:
describe every object, person, animal, text, surface, background, lighting, colors, textures,
spatial relationships, and any other visual details you can observe. Be exhaustive and precise.
\end{llmprompt}

\subsection{Extended Reasoning and Few-Shot}
\label{app:improvement}

\begin{llmprompt}{Extended Reasoning Query (Same Task Prompt)}
What is the primary object that is completely occluded in this image? Be as specific as you can.
\end{llmprompt}

\paragraph{Held-Out Few-Shot Examples.} {\tightpar The few-shot condition prepends two assistant traces to the base query: one direct example and one inferential example. Both items are excluded from the evaluation split for any few-shot run, so few-shot numbers are computed on $378-2=376$ items. Each trace rules out plausible alternatives from context and then justifies its answer using the visible cue.\par}

\begin{llmprompt}{Few-shot Query}
What is the primary object that is completely occluded in this image? Be as specific as you can.
\end{llmprompt}

\begin{llmprompt}{Few-shot Assistant Example 1}
\begin{center}
\includegraphics[width=0.5\textwidth]{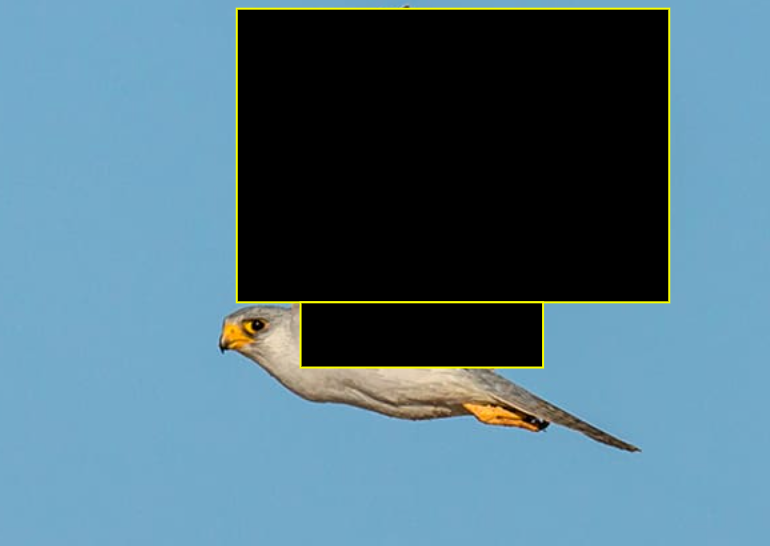}
\end{center}
The answer is not the sky, as the sky is not completely occluded, nor is it the bird,
as the bird is not completely occluded. The only object that is completely occluded is
the bird's wings. It is the bird's wings because, since a bird has wings and it is in
the sky, the only possible object that could be flapping is its wings. Hence the correct
answer is wings.
\end{llmprompt}

\needspace{0.4\textheight}
\begin{llmprompt}{Few-shot Assistant Example 2}
\begin{center}
\includegraphics[width=0.5\textwidth]{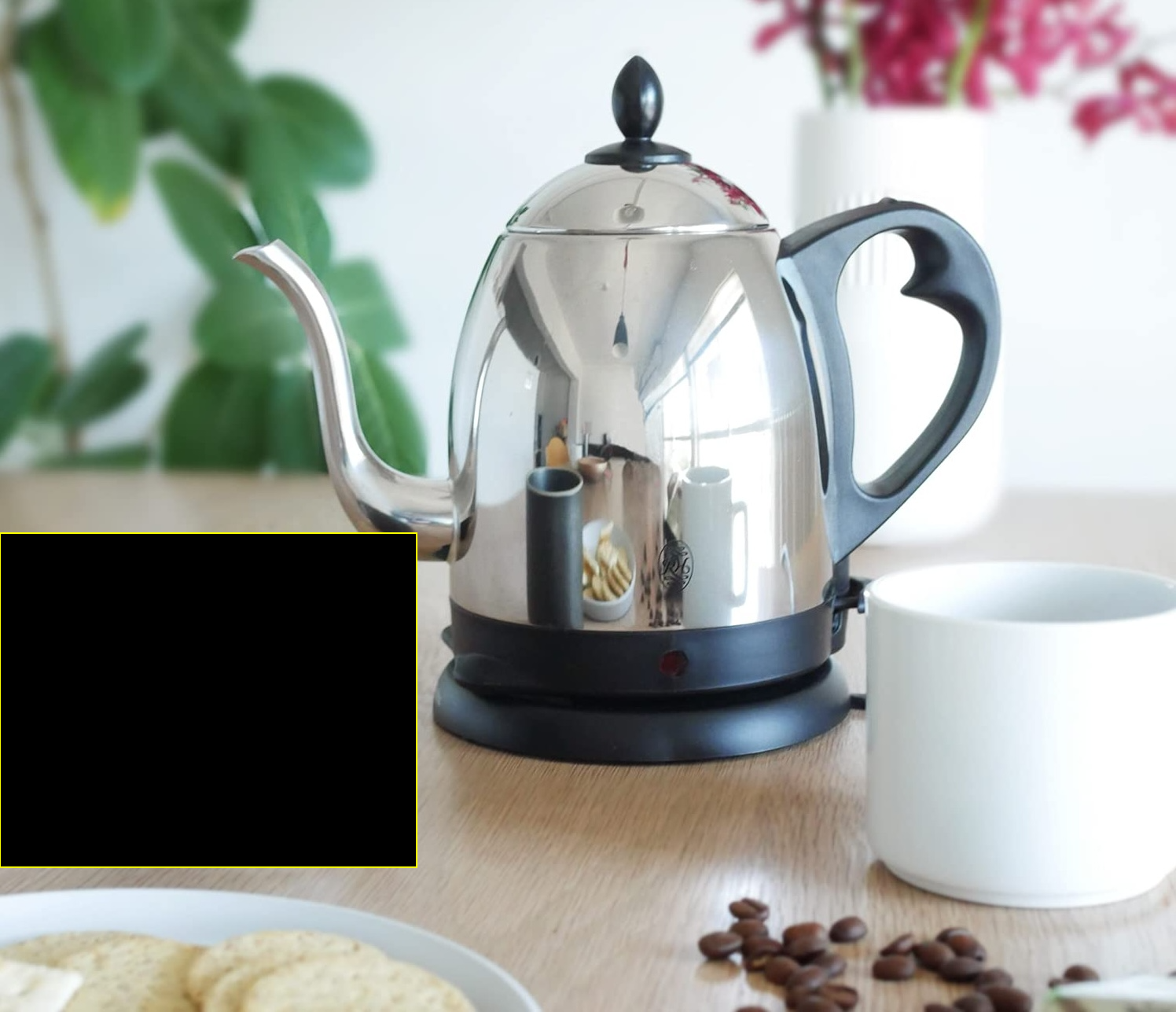}
\end{center}
The answer is not drink, as there are many possible drinks. It's also not just mug, as
there are many different colors of mugs. It is specifically a dark mug. We can arrive at
this conclusion by observing that the teapot is casting a reflection. In the reflection
we observe the white mug and the plate, which are present in the image, but we see that
the dark mug next to the white mug is not present, hence occluded. Thus the answer is
dark mug.
\end{llmprompt}

\subsection{Tool Use}
\label{app:tools}

\paragraph{Tool-Augmented Wrapper Design.} For the tool-use condition, we wrap the same base query inside a per-round controller that gives the model a fixed budget of \texttt{\{MAX\_API\_ROUNDS\}} rounds (configurable; set to 5 in our experiments), requires exactly one tool call per round, and terminates when the model submits a \texttt{final\_answer}. The wrapper passes image dimensions in the prompt so the model can issue valid pixel coordinates directly. Restricting each round to a single tool call prevents the model from issuing many small operations per turn and matches the trace granularity used in the analysis in \secrefc{sec:results} and Appendix~\ref{app:tool_analysis}.

\begin{llmprompt}{Tool-Augmented Wrapper Instruction}
Image dimensions: \{w\}x\{h\} pixels (width x height, origin top-left). You have at most \{MAX\_API\_ROUNDS\} rounds to examine the image. Call exactly one tool per round, then submit a final\_answer when ready.

\{prompt\}
\end{llmprompt}

\begin{llmprompt}{Tool-Augmented Base Query (Occluded)}
What is the primary object that is completely occluded in this image? Be as specific as you can.
\end{llmprompt}

\begin{llmprompt}{Tool Loop Fallback}
Please give your final answer now.
\end{llmprompt}

\paragraph{Tool Palette Rationale.} The tool set covers the basic manipulations a human would try when inspecting a shadow or reflection (zoom, crop, edge detection, flip, rotate), plus a final-answer signal. We exclude higher-level operations such as segmentation, object detection, and OCR, so that gains from tool use reflect geometric inspection rather than offloading recognition to an external module.

\begin{llmprompt}{crop}
    Crops a rectangular region of the current image for closer inspection using pixel coordinates \((x_1, y_1, x_2, y_2)\) with top-left origin. This lets the model isolate a potentially informative local area without altering semantic content.
\end{llmprompt}

\begin{llmprompt}{edge\_detect}
    Applies edge detection to the current image to emphasize boundaries, contours, and high-frequency structure. This is intended to make weak shape cues more visible when texture or shading is ambiguous.
\end{llmprompt}

\begin{llmprompt}{zoom}
    Extracts a rectangular region and rescales it back to the original image dimensions. Unlike simple crop, this preserves the full-frame input size while magnifying a selected area to support fine-grained inspection.
\end{llmprompt}

\begin{llmprompt}{flip\_horizontal}
    Mirrors the current image left-to-right. This provides an alternative orientation that can sometimes make alignment patterns or object asymmetries easier to notice.
\end{llmprompt}

\begin{llmprompt}{flip\_vertical}
    Mirrors the current image top-to-bottom. This offers another geometric transform for inspecting whether weak visual evidence becomes more interpretable under inversion.
\end{llmprompt}

\begin{llmprompt}{rotate}
    Rotates the image counter-clockwise by a specified angle (for example 90, 180, or 270 degrees). This helps when cues are easier to parse in a canonical orientation.
\end{llmprompt}

\begin{llmprompt}{final\_answer}
    Signals termination of tool interaction and submits the model's final predicted object label. This tool does not transform the image; it commits the final textual answer for scoring.
\end{llmprompt}

\subsection{Steering Prompts}
\label{app:nudges}

We test two steering prompts designed to direct VLMs toward indirect visual cues (shadows, reflections, and other physical traces) when answering questions about occluded objects. Both steering prompts are appended to the base prompt used in our main experiments.

\paragraph{Design Intent.} {\tightpar The two prompts vary in strength. The short prompt only points the model at reflective surfaces and shadows and says nothing about valid answers. The long prompt also specifies the answer format (``do not answer `shadow' or `reflection''') and gives the model an inspection order. Comparing them separates gains from attention direction alone from gains that also constrain the answer.\par}

\begin{llmprompt}{Short steering prompt}
Pay close attention to any reflective surfaces or shadows in the image as you will need it to answer the question.
\end{llmprompt}

\begin{llmprompt}{Long steering prompt}
You are answering visual questions about images where a primary object may be completely occluded by a visible mask, box, or polygon.

Your task is to identify the hidden object, not the occluder and not merely a shadow, reflection, surface, or visible nearby object.

Before answering, carefully inspect the whole image. Pay special attention to:
- shadows, reflections, mirrors, glass, water, shiny surfaces, and other indirect visual cues;
- whether those cues physically correspond to the hidden object given position, lighting, orientation, scale, and scene geometry;
- surrounding context, but only as secondary evidence when direct visual cues are insufficient.

Do not rely mainly on common-sense priors if a shadow or reflection provides more specific evidence. Do not answer "shadow" or "reflection" unless the question specifically asks for the visual phenomenon itself; answer the object casting the shadow or being reflected.

Give your best single answer. Be specific enough to distinguish the object when the image supports it, but do not over-specify uncertain details.
\end{llmprompt}

\paragraph{Setup.} We evaluate both steering prompts on all 5 models in the base condition, and additionally GPT-5.4 and Claude Opus 4.6 across thinking and tool-use settings. The long steering prompt is only applied in the thinking and tool-use settings.

\subsection{Inpainting Prompts}
\label{app:inpainting}

\paragraph{Two Stages.} \looseness=-1 The inpainting intervention (\secrefc{subsec:improve}, ``Inpainting Multi-step Execution'') has two stages. In the first, the model is asked to inpaint the occluded region while leaving the rest of the image unchanged; GPT-5.4 uses its native image-generation tool and Gemini 3.1 Pro uses the image-generation variant of the same model family. In the second, a fresh call to the model shows the occluded original together with the generated infill and asks the model to judge the infill's physical consistency before re-answering the identification question. The second-stage answer is scored with the same judges as the main evaluation.

\begin{llmprompt}{Inpainting Stage 1: Fill Occluded Region}
This image has a black region covering something.
Generate a new version of this image with the hidden object naturally visible in place of the black region, keeping everything else in the image exactly the same.
\end{llmprompt}

\begin{llmprompt}{Inpainting Stage 2: Confront and Re-Answer}
The first image is the original: a region of it is covered by a solid mask, hiding an object. The second image is a generated version of that image, where the hidden object has been filled in.

Does this infill make sense physically, given the properties of reflections and shadows? Now, what is the occluded object? Be as specific as you can.
\end{llmprompt}

\FloatBarrier
\section{Performance Across Occlusion Categories}
\label{app:category_uniformity}
\suppressfloats[t]

\begin{figure}[htbp]
    \centering
    \renewcommand{\arraystretch}{1.15}
    \setlength{\abovecaptionskip}{8pt}
    \includegraphics[width=\textwidth]{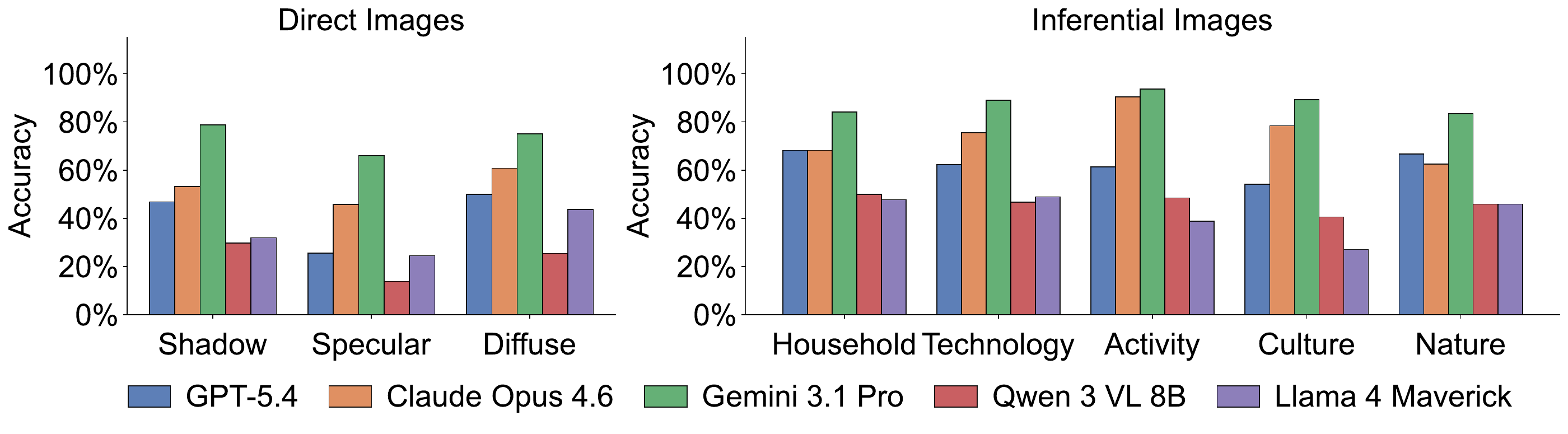}
    \caption{\tightpar Per-category accuracy across occlusion cue categories for each model, separated by image type. Accuracy differences across subcategories within the direct (left) and inferential (right) image categories are mostly not statistically significant, suggesting that semantic differences within the dataset have little effect.}
    \label{fig:category_accuracy}
\end{figure}

\paragraph{Test Design.} \looseness=-1 The category structure of \benchname{} (specular reflection, diffuse reflection, shadow on the direct side; technology, household, culture, activities, nature on the inferential side) raises the natural question of whether the gap between direct and inferential images is really a gap between image \emph{types}, or whether it is driven by one or two unusually difficult subcategories. To rule the latter out, we run an omnibus chi-square test of independence for each model, separately on direct and inferential images, with category as the grouping variable and per-image correctness as the outcome. The null hypothesis is that accuracy is uniform across the cue subcategories within an image type. \figref{fig:category_accuracy} shows per-category accuracy, and \tabref{tab:category_chi2} reports the test results.

\begin{table}[htbp]
\centering
\small
\renewcommand{\arraystretch}{1.15}
\setlength{\abovecaptionskip}{10pt}
\begin{tabular}{lcc}
\toprule
\textbf{Model} & \textbf{Direct $p$} & \textbf{Inferential $p$} \\
\midrule
GPT-5.4              & \textbf{0.0037} & 0.7488 \\
Claude Opus 4.6      & 0.2020 & 0.1242 \\
Gemini 3.1 Pro       & 0.2275 & 0.7190 \\
Qwen 3 VL 8B         & 0.0557 & 0.9393 \\
Llama 4 Maverick     & 0.0524 & 0.2715 \\
\bottomrule
\end{tabular}
\caption{\tightpar Omnibus chi-square $p$-values testing if accuracy varies by cue category. Bold indicates $p < .05$.}
\label{tab:category_chi2}
\end{table}

\paragraph{Takeaway.} {\tightpar The gap between direct and inferential images is not driven by one or two difficult cue subcategories. With accuracy roughly uniform across categories within each image type, the gap reflects differing evidence requirements rather than image difficulty, semantic domain, or annotation noise.\par}

\FloatBarrier
\section{Dataset Breakdown and Sensitivity}
\label{app:dataset-breakdown}
\suppressfloats[t]

\paragraph{Dataset Breakdown and Accuracy.} Out of 378 images, 353 (93.4\%) are collected from the Internet, 19 (5.0\%) come from Visual Genome, and 6 (1.6\%) are taken by the authors. The Visual Genome and author-taken images were included as a small source-diversity check rather than as equally sized comparison groups. Because these two subsets are small, the per-source accuracies in \tabref{tab:source_chi2} are noisy and should be interpreted with caution.

\begin{table}[h]
\centering
\small
\renewcommand{\arraystretch}{1.15}
\setlength{\abovecaptionskip}{10pt}
\begin{tabular}{lcccc}
\toprule
\textbf{Model} & \textbf{Internet} & \textbf{Visual Genome} & \textbf{Author-taken} & $\bm{\chi^2 p}$ \\
\midrule
Gemini 3.1 Pro   & 79\% & 89\% & 67\% & 0.41 \\
Claude Opus 4.6  & 64\% & 63\% & 17\% & 0.061 \\
GPT-5.4          & 50\% & 32\% & 50\% & 0.28 \\
Qwen 3 VL 8B     & 34\% & 16\% & 33\% & 0.26 \\
Llama 4 Maverick & 39\% & 11\% & 0\%  & \textbf{0.008} \\
\bottomrule
\end{tabular}
\caption{Chi-square $p$-values and accuracy across different image sources. Bold indicates $p < .05$.}
\label{tab:source_chi2}
\end{table}

\paragraph{Annotator Disagreement.} After annotating images individually, we reviewed the annotated images and discussed whether the cues in the images are sufficient to identify the occluded object. We ended up discarding 41 images due to the annotator disagreement, which resulted in a 9.8\% rejection rate. We did not have any disagreements on whether the image should be classified as inferential or direct, because it only depends on whether a shadow or reflection is the primary cue in the image.

\paragraph{Sensitivity to Bounding Box Properties.} {\tightpar Annotators unanimously agreed that the occluded mask only occupies the object and has limited blocking of other objects, i.e., no part of the object leaks outside the mask and no other object is fully occluded, and that the manual annotation adds no artifacts.\par}

We change the background color of the polygon as well as the color of the border to see if that affects the accuracy, and find that the accuracy is not sensitive to the changes, as shown in \tabref{tab:box_style_comparison}.

\begin{table}[h]
\centering
\small
\renewcommand{\arraystretch}{1.15}
\setlength{\abovecaptionskip}{10pt}
\begin{tabular}{llcccc}
\toprule
\textbf{Model} & \textbf{Category} & \textbf{n} & \textbf{Original Box} & \textbf{New Box} & $\bm{\Delta}$ \\
\midrule
GPT-5.4          & Overall     & 378 & 49.5\% & 50.0\% & +0.5 \\
GPT-5.4          & Direct      & 197 & 37.6\% & 40.1\% & +2.5 \\
GPT-5.4          & Inferential & 181 & 62.4\% & 60.8\% & $-$1.7 \\
Claude Opus 4.6  & Overall     & 378 & 63.0\% & 60.3\% & $-$2.6 \\
Claude Opus 4.6  & Direct      & 197 & 51.8\% & 46.2\% & $-$5.6 \\
Claude Opus 4.6  & Inferential & 181 & 75.1\% & 75.7\% & +0.6 \\
\bottomrule
\end{tabular}
\caption{Accuracy with the original and new bounding boxes (blue interior, yellow border).}
\label{tab:box_style_comparison}
\end{table}

\FloatBarrier
\section{Difficulty Across Categories}
\label{app:difficulty-distribution}
\suppressfloats[t]

To verify that our benchmark does not conflate category membership with question difficulty, we examine how per-image human accuracy is distributed within the \emph{direct} and \emph{inferential} question sets.

\begin{figure}[htbp]
    \centering
    \includegraphics[width=\linewidth]{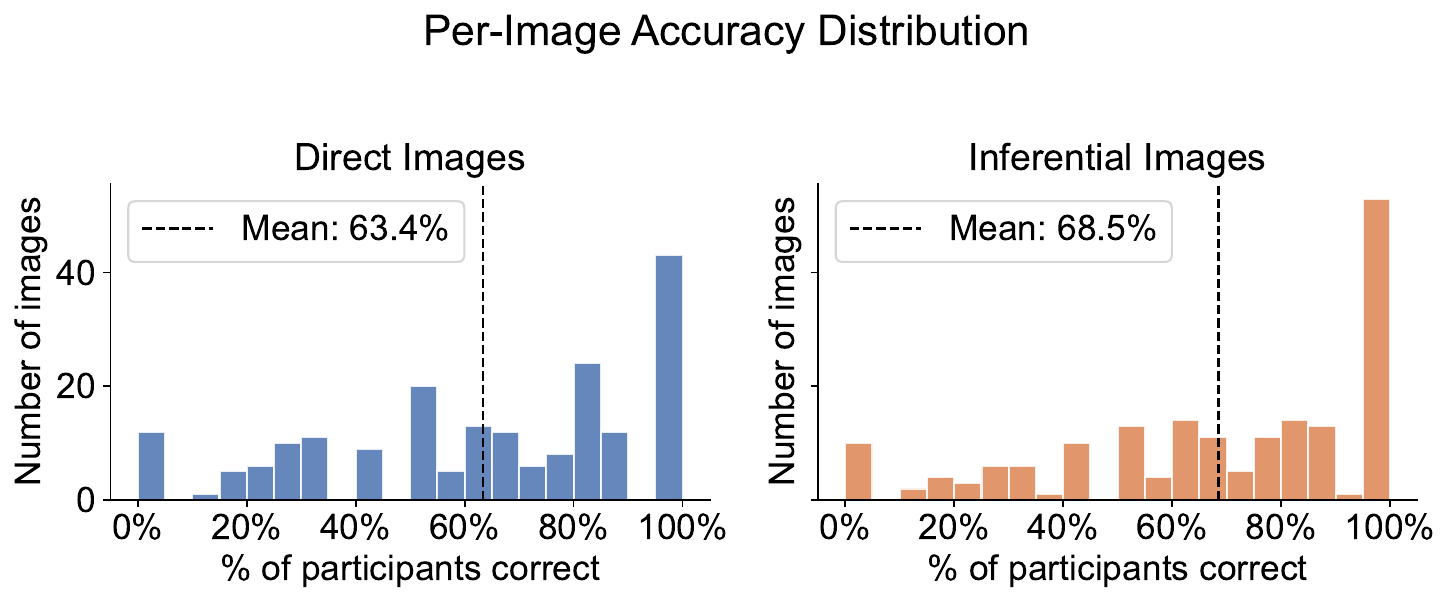}
    \caption{Per-image human accuracy distributions for direct and inferential questions. Both distributions span the full difficulty range, with no concentration at the extremes.}
    \label{fig:per-image-dist}
\end{figure}

\paragraph{Neither category is dominated by polarized difficulty.}
\figref{fig:per-image-dist} shows the distribution of per-image accuracy across all 378 benchmark images (197 direct, 181 inferential), where per-image accuracy is defined as the fraction of Prolific participants who answered that image correctly. Both distributions span the full difficulty range, from images where nearly no participant answered correctly to images where nearly all did, with substantial mass in the middle. This indicates that neither category consists predominantly of trivially easy or impossibly hard questions.

\begin{figure}[htbp]
    \centering
    \includegraphics[width=\linewidth]{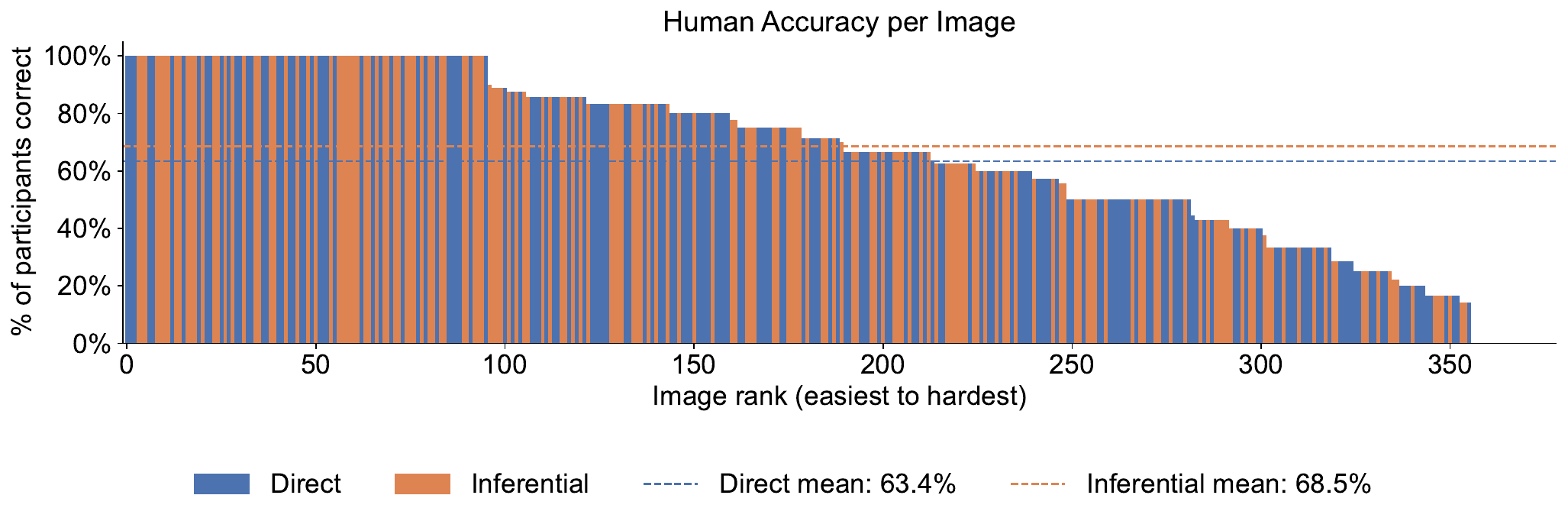}
    \caption{\tightpar Images ranked by human accuracy (easiest to hardest), colored by category. The two categories are interleaved across the difficulty range. Dashed lines show per-category means (direct: 63.4\%, inferential: 68.5\%).}
    \label{fig:per-image-sorted}
\end{figure}

\paragraph{The two categories do not differ significantly in difficulty.}
\looseness=-1 The mean per-image human accuracy is 63.4\% for direct questions and 68.5\% for inferential questions. A Mann-Whitney U test on the per-image accuracy distributions finds no statistically significant difference between the two categories ($U = 15977.5$, $p = 0.079$). \figref{fig:per-image-sorted} further illustrates this: when images are ranked by difficulty, direct and inferential questions are interleaved throughout the difficulty spectrum rather than clustered at opposite ends.

\paragraph{Nearly every image is solvable.} As a check that the benchmark does not rely on intrinsically unsolvable examples, at least one human participant correctly identified the occluded object in 94.2\% of images (356/378), including 93.9\% of direct and 94.5\% of inferential examples.

\paragraph{Lexical Analysis of Direct and Inferential.} {\tightpar We check the lexical prevalence of answer words for direct and inferential questions to test whether one is textually harder, using metrics from \citet{Brysbaert2014Concreteness}. We find no difference in the concreteness of direct and inferential questions (\tabref{tab:lexical_metrics}).\par}

\begin{table}[h]
\centering
\small
\renewcommand{\arraystretch}{1.15}
\setlength{\abovecaptionskip}{10pt}
\begin{tabular}{lcc}
\toprule
\textbf{Metric} & \textbf{Direct} & \textbf{Inferential} \\
\midrule
Zipf freq (Higher = more common) & 4.48 & 4.22 \\
Concreteness (1 = more abstract, 5 = more concrete) & 4.65 & 4.54 \\
Prevalence (fraction of people who know the word) & 1.000 & 0.996 \\
Out-of-norms (lower = fewer rare terms) & 0.035 & 0.067 \\
\bottomrule
\end{tabular}
\caption{Comparison of lexical metrics between direct and inferential categories.}
\label{tab:lexical_metrics}
\end{table}

Taken together, these results indicate that observed performance gaps between direct and inferential questions in model evaluations reflect genuine differences in task type rather than a confound introduced by systematically easier or harder questions in one category.

\FloatBarrier
\section{Tool Use Analysis}
\label{app:tool_analysis}
\suppressfloats[t]

\begin{figure}[htbp]
    \centering
    \includegraphics[width=\textwidth]{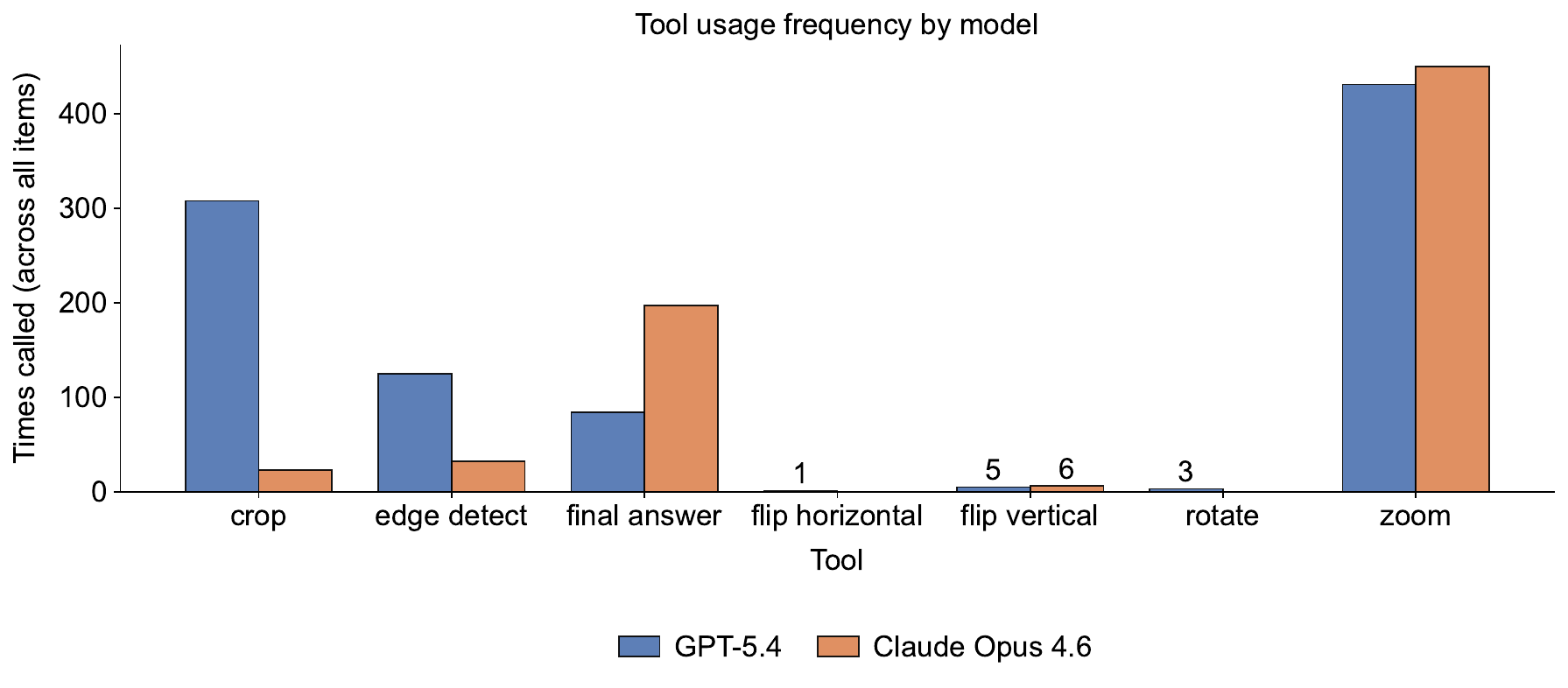}
    \caption{Tool usage frequency by model on direct images.}
    \label{fig:tool_usage_frequency}
\end{figure}

\paragraph{Per-Tool Frequencies.} 
{\tightpar \figref{fig:tool_usage_frequency} shows tool-usage frequency for GPT-5.4 and Claude Opus 4.6 on direct images. GPT-5.4 leans heavily on \texttt{crop} and \texttt{zoom} and repeatedly narrows its view; Claude Opus 4.6 spreads calls more evenly across tools and rarely crops, alternating geometric transforms.\par}

\paragraph{Budget Exhaustion.} GPT-5.4 nearly always exhausts its 5-call budget; Claude Opus 4.6 rarely does. Read together with \figref{fig:tool_call_analysis} in the main text, which shows that Claude Opus 4.6's accuracy \emph{declines} monotonically from 62\% at two calls to 28\% at five, this suggests that additional tool calls reflect indecision rather than progress. For both models, the correct-answer group uses fewer tools on average than the incorrect-answer group.

\paragraph{Implication.} {\tightpar Tool-call volume is a poor proxy for visual grounding effort. The target of a call (whether it lands on the correct projection region) matters more than the count, and even agentic harnesses with unrestricted code execution and high reasoning fail to close the direct--inferential gap.\par}

\FloatBarrier
\section{Details on Human Study}
\suppressfloats[t]

\subsection{Human Study Interface}

\begin{figure}[htbp]
    \centering
    \begin{minipage}[t]{0.68\linewidth}
        \centering
        \includegraphics[height=3.9cm]{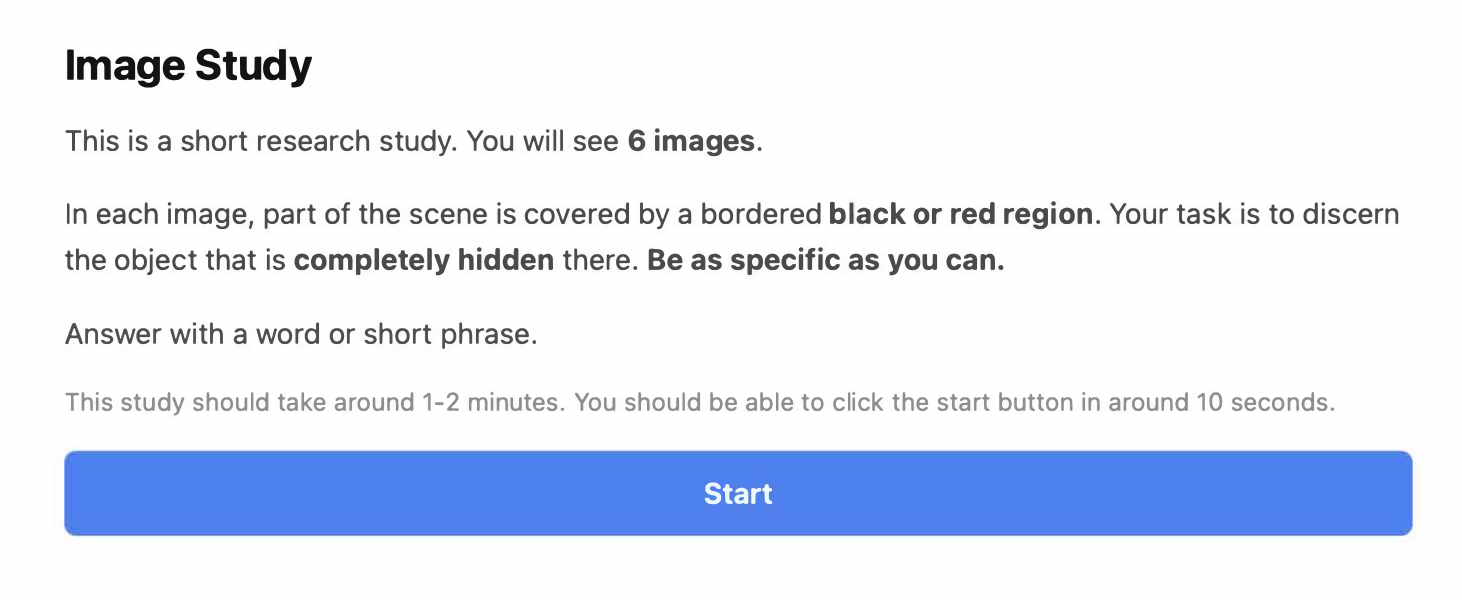}
        \caption{Loading screen of the human study.}
        \label{fig:human_baseline_loading}
    \end{minipage}\hfill
    \begin{minipage}[t]{0.28\linewidth}
        \centering
        \includegraphics[height=3.9cm]{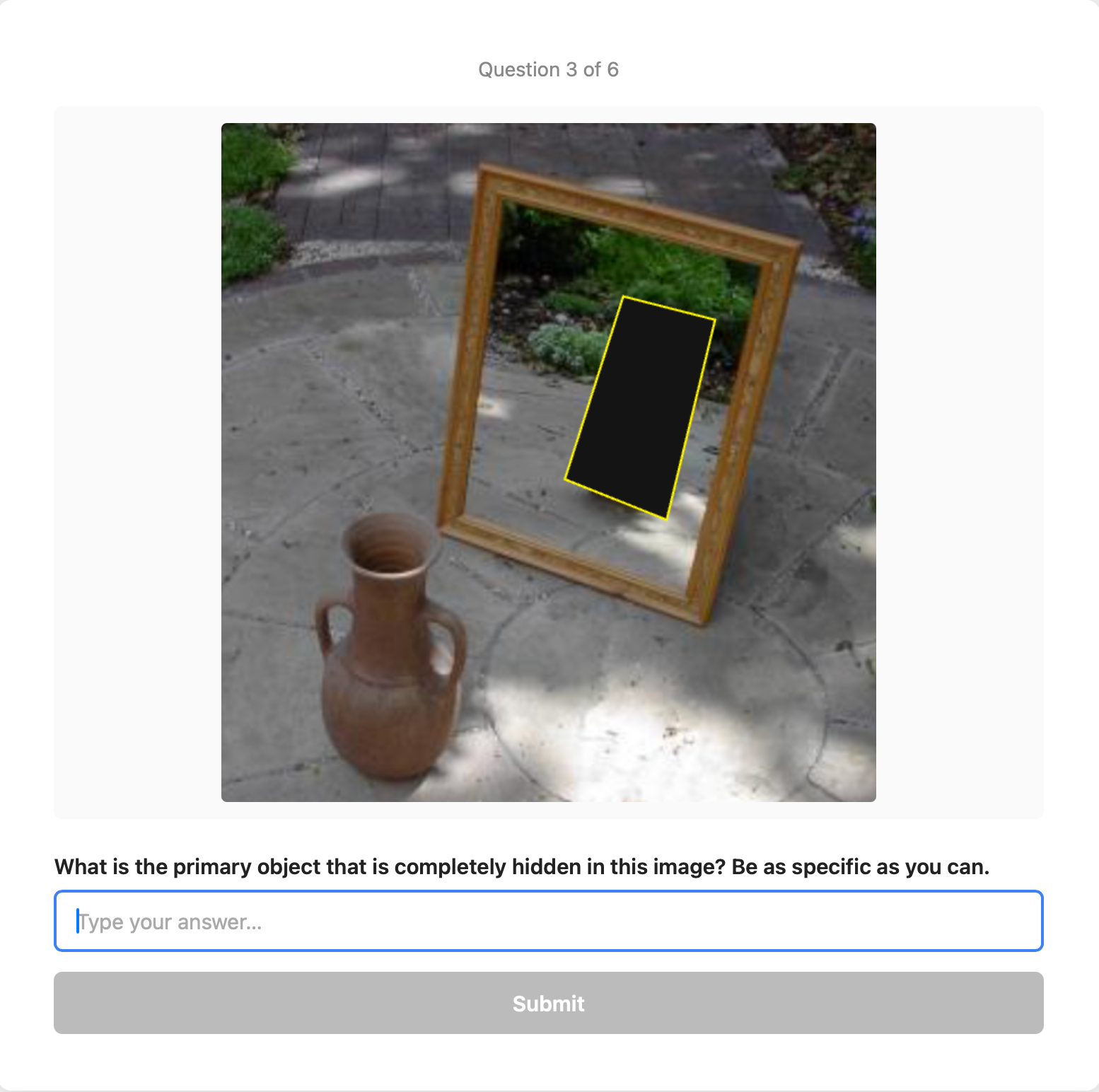}
        \caption{Example question.}
        \label{fig:human_baseline_question}
    \end{minipage}
\end{figure}

This appendix documents the participant-facing interface used to collect the human baseline reported in \secrefc{sec:method}. Figures~\ref{fig:human_baseline_loading} and~\ref{fig:human_baseline_question} show the loading screen and an exemplar question.

\paragraph{Loading Screen.} \figref{fig:human_baseline_loading} shows the screen participants see before the task begins. It is where compensation, study scope, and consent information are first shown.

\paragraph{Per-Question Screen.} {\tightpar \figref{fig:human_baseline_question} shows the layout of a single question. Each participant sees six questions in total: three direct and three inferential, sampled approximately uniformly across the dataset, with item order randomized independently per participant. Items are shown one at a time as an image plus a free-form text box, matching the open-ended response format used for the model evaluations.\par}

\paragraph{Compensation.} Participants are recruited and paid through Prolific~\citep{prolific} at \$12.50/hour, as reported in \secrefc{sec:method}. The median completion time of two minutes thirty seconds is consistent with the per-question design (six short free-form responses) and was used to set the rate.

\subsection{Learning Effects}

{\tightpar We order each participant's six responses by timestamp and regress accuracy and response time on question position. Accuracy is stable across positions (\tabref{tab:q_accuracy}), staying between 62\% and 70\% with no significant increasing/decreasing trend (Cochran-Armitage $p = 0.108$), and response times are flat throughout. Thus, answering multiple questions in the survey does not improve performance.\par}

\begin{table}[htbp]
\centering
\small
\renewcommand{\arraystretch}{1.15}
\setlength{\abovecaptionskip}{10pt}
\begin{tabular}{lcccccc}
\toprule
\textbf{Question position} & 1 & 2 & 3 & 4 & 5 & 6 \\
\midrule
Accuracy (\%) & 69.9 & 65.3 & 64.2 & 65.1 & 67.3 & 62.0 \\
\bottomrule
\end{tabular}
\caption{Human accuracy by question position.}
\label{tab:q_accuracy}
\end{table}

\end{document}